%% file: main.tex
\documentclass{article} % For LaTeX2e
\usepackage{iclr2026_conference,times}
\usepackage{amsmath} % assumes amsmath package installed
\usepackage{amssymb}  % assumes amsmath package installed
\usepackage{mathtools}
\usepackage{graphics}
\usepackage{subcaption}
\input{math_commands.tex}

\usepackage{multirow}
\usepackage{hyperref}
\usepackage{xcolor}
\usepackage{url}
\usepackage{setspace}
\usepackage{enumitem}
\usepackage{booktabs}
\usepackage{wrapfig}
\newcommand{\am}{\ensuremath{\pm}}
\newcommand{\ours}{\texttt{AdaConG}}

\usepackage{amsthm}
\usepackage{algpseudocode}
\usepackage{algorithm}

\hypersetup{
    colorlinks=true,
    linkcolor=black,    
    urlcolor=black,
    citecolor=black
    }

\theoremstyle{plain}

\theoremstyle{definition}

\theoremstyle{remark}

\title{Adaptive Conformal Guidance for Learning under Uncertainty}

\author{
Rui Liu$^{1}$,
Peng Gao$^{2}$,
Yu Shen$^{3}$,
Ming Lin$^{1}$,
Pratap Tokekar$^{1}$, \\
$^{1}$University of Maryland, College Park \\
$^{2}$North Carolina Satate University \\
$^{3}$Adobe Research \\
}

\iclrfinalcopy 
\begin{document}

\maketitle

% \vspace{-10pt}
\begin{abstract}
\vspace{-5pt}
Learning with guidance has proven effective across a wide range of machine learning systems. Guidance may, for example, come from annotated datasets in supervised learning, pseudo-labels in semi-supervised learning, and expert demonstration policies in reinforcement learning. However,  guidance signals can be noisy due to domain shifts and limited data availability and may not generalize well. Blindly trusting such signals when they are noisy, incomplete, or misaligned with the target domain can lead to degraded performance. To address these challenges, we propose \underline{Ada}ptive \underline{Con}formal \underline{G}uidance (\ours), a simple yet effective approach that dynamically modulates the influence of guidance signals based on their associated uncertainty, quantified via split conformal prediction (CP). By adaptively adjusting to guidance uncertainty, \ours~ enables models to reduce reliance on potentially misleading signals and enhance learning performance. We validate \ours~across diverse tasks, including knowledge distillation, semi-supervised image classification, gridworld navigation, and autonomous driving. Experimental results demonstrate that \ours~improves performance and robustness under imperfect guidance, e.g., in gridworld navigation, it accelerates convergence and achieves over $6\times$ higher rewards than the best-performing baseline. These results highlight \ours~as a broadly applicable solution for learning under uncertainty.
\end{abstract}

\vspace{-10pt}
\section{Introduction}
\vspace{-5pt}
Machine learning systems often rely on some form of guidance during training to enhance performance \citep{hinton2015distilling, romero2014fitnets, zagoruyko2016paying, passalis2018learning}, bootstrap learning in data-scarce scenarios \citep{sohn2020fixmatch, zhang2021flexmatch}, and improve sample efficiency \citep{hu2023imitation, bhaskar2024planrl}. While such guidance has proven valuable, a critical challenge arises when this guidance is noisy. In supervised learning, richly annotated datasets provide guidance to enhance model performance, and leveraging pretrained models \citep{hinton2015distilling, romero2014fitnets, zagoruyko2016paying, passalis2018learning, kim2018paraphrasing, wang2023prototype, xue2022modality, huo2024c2kd, gu2023knowledge, jin2023multi} has become an effective strategy to boost performance and enable deployment in resource‑constrained environments, with lighter-weight models that either use reduced modalities or smaller architectures during inference \citep{shen2023auxiliary, liu2025caml, gu2023knowledge, liu2025mmcd}. These teacher–student frameworks allow the student to benefit from the teacher’s superior predictions. However, this setup critically assumes that the teacher’s outputs remain reliable when applied to the student’s target domain. In practice, domain shifts can render the teacher’s guidance noisy or misleading. 

Similarly, semi-supervised learning (SSL) expands the effective training set through pseudo-labeling to bootstrap learning in data-scarce scenarios \citep{sohn2020fixmatch, zhang2021flexmatch}. However, the quality of these pseudo-labels may not be high due to the inherent uncertainty. This uncertainty stems from several factors \citep{scherer2022pseudo, xia2023learning, kage2024review}: the small labeled set may not fully represent the data distribution, the model's early mistakes can propagate through self-training, and the confidence thresholds for pseudo-labeling may not perfectly filter out incorrect labels. As a result, noisy pseudo-labels can misguide the learning process, potentially reinforcing errors and degrading model performance. 

In reinforcement learning, agents employ imitation-learned policies for guidance \citep{hu2023imitation, bhaskar2024planrl} to reduce exploration demands and improve sample efficiency, yet a critical challenge emerges when the target environment differs from the ones used for the expert demonstrations. While imitation learning \citep{liu2024adaptive, liu2025imrl, bhaskar2024lava} provides valuable behavioral priors, these policies often struggle to generalize beyond their training distribution. When the RL agent encounters states or observations outside the expert's demonstration space, the imitation policy's guidance becomes increasingly noisy, potentially leading to suboptimal exploration. 

\begin{figure}[t]
    \centering
    \includegraphics[width=0.95\linewidth]{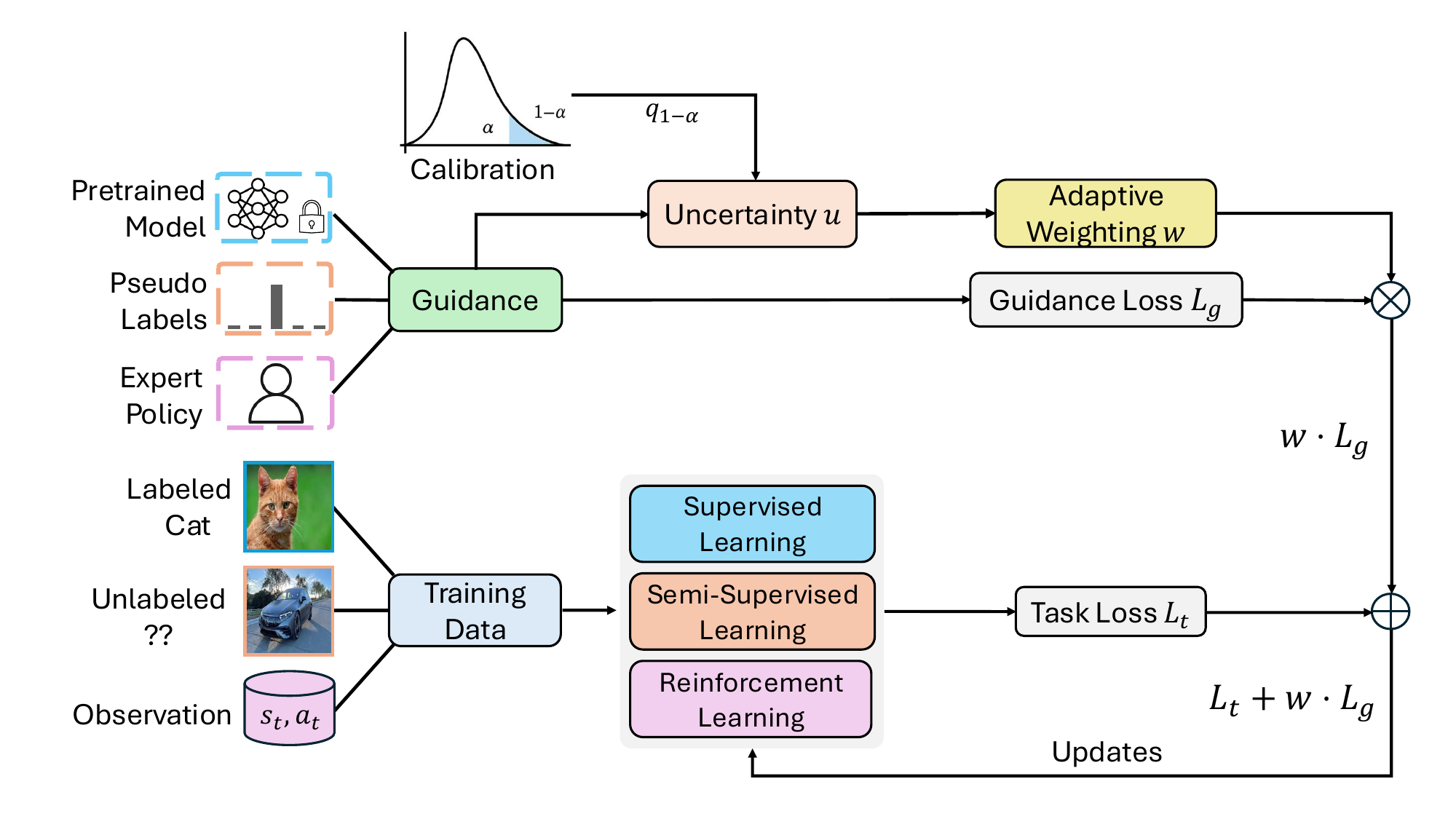}
    \vspace{-10pt}
    \caption{\textbf{Overview of the \ours~approach}. \ours~leverages split CP with calibration to quantify the uncertainty of guidance signals and adaptively modulate their influence. The estimated uncertainty $u$ is converted into an adaptive weight $w$, which reweights the guidance loss. This weighted guidance loss is then combined with the task loss to update the model, enabling effective learning under uncertain guidance.}
    \label{fig:app}
    \vspace{-16pt}
\end{figure}

In all these scenarios, blindly relying on noisy guidance can propagate errors and lead to suboptimal model performance, as the learning system overfits to potentially misleading information, yet discarding potentially valuable guidance wastes computational resources and domain knowledge. Despite its fundamental importance, the challenge of effectively leveraging noisy guidance while maintaining robust learning capabilities remains largely unaddressed across machine learning systems. The central question becomes: \textit{How can we effectively leverage potentially valuable guidance while appropriately accounting for its uncertainty to ensure robust model learning?}

While prior works have explored uncertainty‑aware learning \citep{angelopoulos2020uncertainty, mossina2024conformal, lu2022improving, karimi2023quantifying, zhao2024leveraging, su2025eakdentropybasedadaptiveknowledge, zhang2024knowledge, edupuganti2020uncertainty, kwon2020uncertainty, wang2020double, gao2021bayesian, gao2023uncertainty, liu2023data, liu2024towards}, they typically focus on heuristic uncertainty estimates, domain‑specific solutions, or post‑hoc calibration. Monte Carlo dropout \citep{zhao2024leveraging}, entropy‑based reweighting \citep{su2025eakdentropybasedadaptiveknowledge}, or perturbation of the distillation loss \cite{zhang2024knowledge}, tackle uncertainty by modulating loss terms via heuristics or series approximations. However, these approaches remain confined to supervised classification, and the effectiveness may degrade under distribution shifts. Other methods address uncertainty only within narrow domains, e.g., medical imaging \citep{edupuganti2020uncertainty, kwon2020uncertainty, wang2020double} or human-robot interaction \cite{gao2021bayesian, gao2023uncertainty}. Techniques such as conformal prediction (CP) have been used primarily to calibrate model outputs after training \citep{angelopoulos2020uncertainty, mossina2024conformal, lu2022improving, karimi2023quantifying}.

% \begin{wrapfigure}{r}{0.5\textwidth} % r for right, l for left
% \vspace{-13pt}
%     \centering
%     \includegraphics[width=\linewidth]{figs/fig11.png}
%     \caption{\textbf{\ours~framework.} Our framework adaptively modulates the influence of guidance signals based on their associated uncertainty. When guidance exhibits high confidence, the model prioritizes this external knowledge; conversely, when uncertainty is high, the model reduces its reliance on potentially misleading guidance and explores the data more independently.}
%     \label{fig:moti}
% \vspace{-15pt}
% \end{wrapfigure}

In contrast, we propose the first framework \ours~that embeds split CP directly into the training loop to adaptively weight guidance across supervised, semi‑supervised, and reinforcement learning settings. Simple yet effective, our method serves as a broadly applicable solution for incorporating uncertainty‑aware guidance, as illustrated in Fig. \ref{fig:app}. While CP has primarily been applied for post-hoc calibration, its potential to inform real-time training dynamics remains underexplored. CP provides a distribution-free, model-agnostic approach to constructing prediction sets \citep{shafer2008tutorial, angelopoulos2021gentle}, making it suited across diverse learning systems. Unlike heuristic uncertainty estimates such as entropy \citep{namdari2019review} or maximum softmax probability (MSP) \citep{pearce2021understanding} which rely on the softmax outputs that are often overconfident and poorly calibrated, CP provides more rigorous uncertainty estimates \citep{shafer2008tutorial, angelopoulos2021gentle}, even when the underlying distribution changes \citep{zhou2025computation, gibbs2021adaptive}.

We validate our approach through extensive experiments across multiple tasks, including knowledge distillation, semi-supervised image classification, gridworld navigation, and autonomous driving, demonstrating improvements in performance and robustness compared to conventional methods. Our results underscore the critical importance of adaptive uncertainty weighting in scenarios where guidance signals may be imperfect, providing a solution toward more reliable machine learning systems. Overall, the key contributions of our work are as follows:
\vspace{-6pt}
\begin{itemize}[left=0pt]
    \item We propose \ours, an approach that adaptively modulates the influence of guidance signals based on their  uncertainty, ensuring effective learning without over-relying on unreliable guidance.
    
    \vspace{-6pt}
    \item \ours~is broadly applicable across diverse learning systems including supervised, semi-supervised, and imitation-guided reinforcement learning.
    \vspace{-6pt}
    \item \ours~can extract useful insights even when guidance underperforms, unlike conventional methods that assume guidance is always trustworthy. In gridworld navigation, it enables faster convergence and achieves over $6\times$ higher rewards than the strongest baseline.
\end{itemize}

\vspace{-10pt}
\section{Related Work}
\vspace{-5pt}
\paragraph{Learning with Guidance.}
Learning with guidance has been a common and effective strategy across various machine learning systems. In supervised learning, annotated datasets provide explicit guidance for model training, and many works leverage pretrained models to further boost performance. For instance, \citet{hinton2015distilling, jin2023multi, sun2024logit} focused on transferring soft probabilities from teacher models' logits to guide student models, while \citet{romero2014fitnets, zagoruyko2016paying, passalis2018learning, kim2018paraphrasing} emphasized transferring intermediate features. Cross-modal guidance has also been explored: \citet{wang2023prototype} proposed a prototype-based distillation method for medical image segmentation, where a multi-modal teacher guides a single-modal student; \citet{shen2023auxiliary} introduced the Auxiliary Modality Learning (AML) framework, enabling a teacher model with access to multiple modalities to transfer knowledge to a student operating with fewer modalities at test time; and \citet{liu2025caml} extended this idea to multi-agent settings. In semi-supervised learning \citep{sohn2020fixmatch, zhang2021flexmatch}, pseudo-labels generated from unlabeled data provide implicit guidance to bootstrap learning with limited labeled data. In reinforcement learning, pretrained imitation learning (IL) policies \citep{hu2023imitation, bhaskar2024planrl} derived from expert demonstrations have been used to guide RL agents and improve sample efficiency.

However, a key limitation of most existing methods is their reliance on static guidance, which assumes that guidance signals are always reliable. This assumption often breaks down when guidance contains uncertainty, due to domain shifts in supervised learning, limited labeled data in semi-supervised learning, or generalization constraints of IL policies in reinforcement learning. In contrast, \ours~introduces a principled approach that dynamically modulates the influence of guidance signals based on their associated uncertainty, offering a simple yet effective, and broadly applicable solution for incorporating uncertainty-aware guidance.

\vspace{-6pt}
\paragraph{Conformal Prediction.}
Conformal prediction (CP) \citep{angelopoulos2020uncertainty, angelopoulos2021gentle, mossina2024conformal, karimi2023quantifying, tibshirani2019conformal, shafer2008tutorial, vovk2020conformal} is a non-parametric, distribution-free, and model-agnostic framework designed to provide reliable prediction sets. In machine learning systems, CP has primarily been utilized for post-hoc uncertainty calibration. For instance, \citet{angelopoulos2020uncertainty} introduced an algorithm that adapts any image classifier to output predictive sets containing the true label with a user-specified probability. \citet{mossina2024conformal} proposed a computationally lightweight approach to quantify predictive uncertainty in semantic image segmentation using CP. Similarly, \citet{lu2022improving} applied CP to deep learning models for grading the severity of spinal stenosis in lumbar spine MRI, while \citet{karimi2023quantifying} leveraged CP to measure uncertainty in deep learning models.

Despite recent advancements, the application of CP to inform real-time training dynamics remains underexplored. In this work, we extend split CP to learning with guidance under uncertainty, using it as a module for adaptive weighting. By modulating the uncertainty of the guidance signal, we enable the model to reduce dependence on potentially misleading guidance and encourages the model to discover patterns that may be overlooked when strictly following uncertain guidance.

\section{Approach} \label{sec:app}
\vspace{-6pt}
\paragraph{Preliminaries.} \label{sec:pre}
In our framework, we leverage split conformal prediction (CP) to conformalize a guidance signal and quantify its uncertainty. CP is a distribution-free method that provides prediction sets with guaranteed coverage levels, regardless of the underlying model or data distribution \citep{angelopoulos2021gentle, shafer2008tutorial}.

Split CP uses a nonconformity score $s$ to measure how unusual a prediction is for a new test input, based on a calibration set $\mathcal{D}_{\text{cal}}$, a held-out dataset used to compute the empirical distribution of nonconformity scores. The score $s$ can be defined in various ways. For instance, in regression, it is often the absolute residual $s = |\bar{y} - \hat{y}(\bar{x})|$, where $\bar{y}$ is the ground truth and $\hat{y}(\bar{x})$ is the model's prediction for an input $\bar{x} \in \mathcal{D}_{\text{cal}}$. In classification, a common choice is the confidence score $s = 1 - p_{\bar{y}}(\bar{x})$, where $p_{\bar{y}}(\bar{x})$ is the model's estimated probability for the true class $\bar{y}$. Additional examples can be found in \citep{angelopoulos2021gentle, shafer2008tutorial}. Given a calibration set, we compute the quantile $q_{1-\alpha}$ of the nonconformity scores with $\alpha$ as an allowable error rate. The quantile is denoted as $q_{1-\alpha} = \text{Quantile}_{1-\alpha}(s_1, s_2, \ldots, s_{|\mathcal{D}_{\text{cal}}|})$, representing a threshold below which $1-\alpha$ of the data falls. This threshold is then used to construct prediction sets. For a test input $x_{\text{test}}$, the prediction set is constructed as $\mathcal{C}(x_{\text{test}}) = \{y: s(x_{\text{test}}, y) \leq q_{1-\alpha}\}$. Under the assumption of exchangeability, the coverage guarantee holds that the probability of the true label $y_\text{test}$ falling within $\mathcal{C}(x_{\text{test}})$ satisfies $P(y_{\text{test}}\in \mathcal{C}(x_{\text{test}})) \geq 1-\alpha$ \citep{angelopoulos2021gentle}.

% ensuring a coverage probability of $1-\alpha$, assuming the calibration and test data are from the same distribution

% \vspace{-8pt}
% \subsection{Problem Definition}
% \vspace{-5pt}
% We consider a machine learning problem with some form of guidance during training. Formally, let the source domain data be denoted as $\mathcal{D}_s$, and the target domain data as $\mathcal{D}_t$. We represent the model pretrained on $\mathcal{D}_s$ as $f_p: \mathcal{X} \rightarrow \mathcal{Y}$, and the target model as $f_t: \mathcal{X} \rightarrow \mathcal{Y}$, where $\mathcal{X}$ is the input space and $\mathcal{Y}$ is the output space. In this framework, we do not retrain or finetune the pretrained model $f_p$ on the target domain $\mathcal{D}_t$, as this can be computationally prohibitive. Instead, we aim to transfer the knowledge from $f_p$ to $f_t$, with the goal of improving the performance of $f_t$ on $\mathcal{D}_t$ compared to conventional knowledge transfer approaches or training the student model from scratch. This is particularly important when $\mathcal{D}_t$ differs from $\mathcal{D}_s$ (i.e., $P_t\neq P_s$), making $f_p(x)$ uncertain for input $x\in \mathcal{D}_t$.

\vspace{-5pt}
\paragraph{Adaptive Conformal Guidance.}
\ours~is a general framework, as illustrated in Fig. \ref{fig:app}. We show how to use it in supervised, semi-supervised, and imitation-guided reinforcement learning settings. The core idea is to learn adaptive weights based on the uncertainty of the guidance signal, enabling dynamic modulation of its influence during training.

\vspace{-5pt}
\subparagraph{Supervised Learning.} \label{sec:sup}
We consider a supervised learning problem in which a pretrained model guides the training of a target model under potential domain shift. Formally, let the source domain dataset be denoted as $\mathcal{D}_s$, and the (shifted) target domain dataset as $\mathcal{D}_t$. We represent the pretrained model on $\mathcal{D}_s$ as $f_p: \mathcal{X} \rightarrow \mathcal{Y}$, and the target model under training as $f_t: \mathcal{X} \rightarrow \mathcal{Y}$, where $\mathcal{X}$ is the input space and $\mathcal{Y}$ is the output space. Our goal is to leverage $f_p$ to bootstrap the learning of $f_t$ so that it outperforms both supervised training from scratch and standard knowledge distillation under domain shift. To do so, we introduce an adaptive weighting mechanism based on split CP to modulate the guidance of the pretrained model. This is particularly important when $\mathcal{D}_t$ differs from $\mathcal{D}_s$, making $f_p(x)$ uncertain for input $x\in \mathcal{D}_t$.

% \begin{wrapfigure}{r}{0.5\textwidth} % r for right, l for left
% \vspace{-28pt}
%     \centering
%     \includegraphics[width=\linewidth]{figs/app.png}
%     \caption{\textbf{\ours~Approach Pipeline.} The figure demonstrates our approach  dynamically adjusts the teacher's guidance based on its prediction uncertainty using conformal prediction. This method balances the exploitation of teacher knowledge with student exploration itself, enabling better transferability of knowledge under domain shifts.}
%     \label{fig:app}
% % \vspace{-10pt}
% \end{wrapfigure}

Given the target domain dataset $\mathcal{D}_t$, we split it into three subsets: the training set $\mathcal{D}_\text{train}^t$, used to train the target model $f_t$; the calibration set $\mathcal{D}_\text{cal}$, used to transform any heuristic measure of uncertainty from the pretrained model $f_p$ into a rigorous one; and the testing set $\mathcal{D}_\text{test}$ for validate the target model performance. This setup ensures that the calibration set is representative of the inputs on which the guidance will be applied. We conformalize the pretrained model $f_p$ following the approach as described in Section \ref{sec:pre}. To quantify the guidance uncertainty, we leverage the size of the prediction set $\mathcal{C}(x)$ for an input $x$. Specifically, we define the guidance uncertainty as 
\begin{equation}
\label{eq: ut}
    u(x) = g(|\mathcal{C}(x)|),
\end{equation}
where $g$ is a mapping ensuring $u(x)\in[0,1]$ (e.g., $g(n)=\frac{n-1}{K-1}$ for a $K$-class problem). The adaptive weight is computed as 
\begin{equation}
\label{eq: weight}
w(x) = h\big(u(x)\big),
\end{equation}
where $h$ is a monotonically decreasing function (e.g., exponential decay $h(u)=\exp(-\gamma u)$ with a temperature $\gamma>0$) such that high uncertainty results in a lower weight. Then we define the loss function of training the target model as
$ \label{eq: loss}
    \gL = \lambda_{\text{task}} \gL_t + w(x)\cdot\lambda_{\text{guide}} \gL_g,
$
where $\gL_t$ is the task loss (e.g., cross entropy loss), $\gL_g$ is the guidance loss (e.g., KL divergence between target model and pretrained model logits), $\lambda_\text{task}$ and $\lambda_\text{guide}$ are the coefficients. Essentially, the adaptive weighting mechanism of \ours~allows the model to balance between relying on the pretrained guidance and self-exploration. 
 
\vspace{-6pt}
\subparagraph{Semi-Supervised Learning.} \label{sec:ssl}
In semi-supervised learning (SSL), our goal is to leverage a limited set of labeled data $\mathcal{D}_l = \{(x_i, y_i)\}_{i=1}^{N_l}$ along with a large set of unlabeled data $\mathcal{D}_u = \{x_i\}_{i=1}^{N_u}$. For each unlabeled sample $x \in \mathcal{D}_u$, the model $f$ produces a prediction $\hat{y} = f(x_{\text{weak}})$ using a weakly augmented view $x_{\text{weak}}$. The pseudo-label $\tilde{y}$ is is obtained by thresholding the model’s output on this weakly augmented input. The same model is then trained to predict $\tilde{y}$ from a strongly augmented version $x_{\text{strong}}$ of the same input. 

Along with obtaining pseudo-labels, we introduce an adaptive weighting mechanism grounded in split CP to modulate the influence of each pseudo-label during training. We construct a calibration set $\gD_{\text{cal}}$ by taking the labeled data and applying the same weak augmentation used for the unlabeled data. The CP procedure for generating a prediction set $\gC(x)$ for an unlabeled sample $x$ follows the approach detailed in Section \ref{sec:pre}. The adaptive weight $w(x)$ is defined similarly as in Eq. \ref{eq: weight}. Then the unsupervised loss is defined as
$
\mathcal{L}_u = \frac{1}{|\mathcal{D}_u|}\sum_{x \in \mathcal{D}_u} w(x)\,\ell\big( f(x_{\text{strong}}), \tilde{y} \big),
$
where $\ell(\cdot,\cdot)$ denotes the cross-entropy loss commonly used in SSL, which enforces consistency regularization between the strongly augmented prediction and the pseudo-label. This loss is adaptively weighted by the confidence of the pseudo-label $w(x)$. The supervised loss over the labeled set is given by another cross-entropy loss
$
\mathcal{L}_s = \frac{1}{|\mathcal{D}_l|}\sum_{(x,y) \in \mathcal{D}_l} \ell\big( f(x), y \big).
$ The final objective function, which balances the contributions from both supervised and unsupervised components, is defined as $\mathcal{L} = \mathcal{L}_s + \lambda_u\mathcal{L}_u,$
% \begin{equation}
% \mathcal{L} = \mathcal{L}_s + \lambda_u\mathcal{L}_u,
% \end{equation}
with $\lambda_u$ controlling the relative weight of the unsupervised loss.

\vspace{-8pt}
\subparagraph{Imitation-Guided Reinforcement Learning.} \label{sec:rl}
Consider a Markov Decision Process defined by the tuple $\{\gS, \gA, \gP, \gR, \gamma\}$, where $\gS$ is the state space, $\gA$ is the action space, $\gP$ is the transition dynamics, $\gR$ is the reward function, and $\gamma$ is the discount factor. We focus on off-policy RL methods due to their higher sample efficiency. We focus on cases where there is an imitation policy learned from expert demonstrations to guide the reinforcement learning. 

To quantify uncertainties of IL and RL policy via split CP, we use the nonconformity score $s(s,a) = -\log \pi(a|s)$. For the static imitation policy $\pi_\text{I}$, we pre-collect a calibration set $\mathcal{D}_{\text{cal, I}}=\{(s_i,a_i)\}_{i=1}^N$ by rolling out $\pi_\text{I}$ in the target environment. Then we compute a constant quantile $\hat{q}_{\text{I}}$ for the IL policy. For the  RL policy $\pi_\text{R}^{(t)}$, we leverage adaptive CP \citep{zhou2025computation, gibbs2021adaptive}, maintaining a dynamic calibration set $\mathcal{D}_{\text{cal, R}}^{(t)}$ via a sliding window of size $N$, which we initialize as $\mathcal{D}_{\text{cal, R}}^{(0)}=\mathcal{D}_{\text{cal, I}}$. At each subsequent training step $t$, we add a new batch of $m$ state-action pairs from rollouts of $\pi_\text{R}^{(t)}$ and discard the oldest. Then we update the RL quantile $\hat{q}_{\text{R}}^{(t)}$ using an exponential moving average (EMA): $\hat{q}_{\text{R}}^{(t)} \leftarrow (1-\gamma)\hat{q}_{\text{R}}^{(t-1)} + \gamma \tilde{q}_{\text{R}}^{(t)}$, where $\tilde{q}_{\text{R}}^{(t)}$ is computed from the current window's scores and $\gamma$ is a smoothing factor. We warm-start the RL quantile by initializing it with the quantile of the imitation policy, i.e., $\hat{q}_{\text{R}}^{(0)} = \hat{q}_{\text{I}}$. Finally, the IL and RL policy uncertainties are defined as $u_\text{I}(s) = g(|\mathcal{C}_\text{I}(s)|)$ and $u_\text{R}(s) = g(|\mathcal{C}_\text{R}(s)|)$, using the static quantile $\hat{q}_{\text{I}}$ for $\pi_\text{I}$ and the adaptive quantile $\hat{q}_{\text{R}}^{(t)}$ for $\pi_\text{R}^{(t)}$, where $g$ is a mapping, e.g., the identity function. The loss for training the RL policy is defined as $\gL = \gL_t + w(s)\cdot \gL_g$, where $\gL_t$ is the task loss (e.g., $\mathbb{E}\left[ -\log f_\text{S}(a|s) \cdot A(s, a)\right]$ with $A(s, a)$ as the advantage function), which corresponds to the RL objective. $\gL_g = \E\left[\text{KL}(\pi_\text{R}(\cdot|s) \parallel \pi_\text{I}(\cdot|s)) \right]$ is the KL divergence between the RL policy and the IL policy, serving as the guidance loss. The weight $w$ is defined as: $w(s) = h(u_\text{I}(s), u_\text{R}(s))$, e.g., $w(s)=\frac{\exp(-u_\text{I}(s))}{\exp(-u_\text{I}(s)) + \exp(-u_\text{R}(s))}$. This reduces reliance on imitation guidance when the IL policy's uncertainty is high. 

\vspace{-6pt}
\section{Experiments} \label{sec:exp}
\vspace{-5pt}
To validate our approach, we conduct experiments across a diverse range of tasks, including knowledge distillation, semi-supervised image classification, gridworld navigation, and autonomous driving steer prediction.

\vspace{-5pt}
\paragraph{Knowledge Distillation.}
We first evaluate our framework's effectiveness in improving classification performance over traditional supervised learning by leveraging knowledge distillation from a pretrained teacher to a student model. This evaluation is conducted under domain shift and noise, where the teacher may underperform. Specifically, we aim to address the following question: When the teacher model is underperformance, can it still provide useful ``dark knowledge'' to enhance the performance of the student model beyond what is achievable by training from scratch?

\vspace{-5pt}
\subparagraph{Experimental Setup.}
We conduct our experiments on the CIFAR-100 dataset \citep{krizhevsky2009learning} and report the mean and standard deviation over four repeated runs. We introduce domain shifts to the datasets by adding Gaussian noise of zero mean and a standard deviation of 0.05, which may lead to underperformance of the teacher model. We evaluate two settings: (1) Homogeneous Structure, where the teacher and student share the same architecture type (e.g., ResNet-32x4 and ResNet-8x4), and (2) Heterogeneous Structure, where the teacher and student use different architectures (e.g., ResNet-32x4 and ShuffleNet-V1). For further details of the models evaluated and the dataset, please refer to Appendix \ref{app:img}. To quantify the prediction uncertainty of the pretrained teacher model $f_p$, we utilize the RAPS algorithm \citep{angelopoulos2020uncertainty}. Given an input image $x$, we obtain the prediction set $\mathcal{C}(x)$ with $\alpha=0.1$ and define the uncertainty as $u(x) = \frac{|\mathcal{C}(x)|-1}{K-1}$ \citep{vovk2016criteria}, where $K=100$ is the total number of classes. The adaptive weight is computed as $w=\exp(-\gamma u)$ with $\gamma=10.0$, as described in Section \ref{sec:sup}. Please see Appendix~\ref{hyper_sp} for an detailed analysis of the design choices for the hyperparameters. We follow the same experimental settings as in previous work \citep{tian2019contrastive, sun2024logit} for the coefficients $\lambda_{\text{task}}$ and $\lambda_{\text{guide}}$, as well as other training details.

\vspace{-5pt}
\subparagraph{Baselines.} 
We measure Top‑1 classification accuracy for a range of baselines, including classic distillation methods, KD \citep{hinton2015distilling}, FitNet \citep{romero2014fitnets}, PKT \citep{passalis2018learning}, FT \citep{kim2018paraphrasing}, and LS-KD \citep{sun2024logit}, evaluating each both with and without our approach \ours, alongside uncertainty‑aware adaptations such as EA‑KD \citep{su2025eakdentropybasedadaptiveknowledge} and PTLoss \citep{zhang2024knowledge}, as well as KD leveraging heuristic confidence estimators including maximum softmax probability (MSP) \citep{pearce2021understanding}, Monte Carlo (MC) dropout \citep{gal2016dropout}, and output entropy \citep{namdari2019review}.

\vspace{-6pt}
\begin{table*}[h]
\caption{Top-1 accuracy (\%) of various knowledge distillation methods on CIFAR-100 under homogeneous structure where the teacher models are underperforming due to domain shift. We use $\Delta$ to show mean performance gain relative to conventional knowledge distillation methods without \ours. Following the protocol in \citep{sun2024logit}, we highlight in \textcolor{orange}{orange} $\Delta$ greater than 0.15, indicating non-trivial enhancement. We observe up to {\bf +10.89\%} higher accuracy.}
\centering
\resizebox{\textwidth}{!}{
\vspace{-5pt}
\begin{tabular}{lcccccc}
\hline \multirow{2}{*}{Teacher} & ResNet110 & ResNet56 & ResNet32 $\times 4$ & VGG13 & WRN-40-2 & WRN-40-2 \\
& 58.78 & 56.23 & 62.61 & 61.47 & 58.76 & 58.76 \\
Student & ResNet20 & ResNet20 & ResNet8 $\times 4$ & VGG8 & WRN-40-1 & WRN-16-2 \\
& 66.51\am0.14 & 66.51\am0.14 & 69.14\am0.21 & 67.18\am0.18 & 69.03\am0.21 & 70.34\am0.20 \\
\hline 
KD \citep{hinton2015distilling} & 57.23\am 0.24 & 56.27\am 0.17 & 58.90\am 0.31 & 61.00\am 0.25 & 58.44\am 0.16 & 59.40\am 0.33 \\
KD + \ours & 66.53\am 0.55 & 66.98\am 0.25 & 68.45\am 0.29 & 67.53\am 0.18 & 69.31\am 0.25 & 70.29\am 0.39 \\
$\Delta$ & \color{orange}{9.30} & \color{orange}{10.71} & \color{orange}{9.45} & \color{orange}{6.53} & \color{orange}{10.87} & \color{orange}{10.89} \\
\hline
FitNet \citep{romero2014fitnets} & 64.65\am0.30 & 64.98\am0.16 & 69.21\am0.17 & 67.19\am0.39 & 68.74\am0.24 & 70.49\am0.27 \\
FitNet + \ours & 67.06\am0.13 & 66.91\am0.14 & 69.49\am0.18 & 67.58\am0.30 & 69.11\am0.18 & 71.00\am0.23 \\
$\Delta$ & \color{orange}{2.41} & \color{orange}{1.93} & \color{orange}{0.28} & \color{orange}{0.39} & \color{orange}{0.37} & \color{orange}{0.51}\\
\hline
PKT \citep{passalis2018learning} & 66.67\am0.17 & 66.54\am0.26 & 69.69\am0.34 & 67.06\am0.09 & 69.12\am0.22 & 70.55\am0.26 \\
PKT + \ours & \textbf{67.55}\am0.11 & \textbf{67.42}\am0.51 & 70.27\am0.39 & 68.50\am0.13 & 70.03\am0.51 & 71.22\am0.47 \\
$\Delta$ & \color{orange}{0.88} & \color{orange}{0.88} & \color{orange}{0.58} & \color{orange}{1.44} & \color{orange}{0.91} & \color{orange}{0.67} \\
\hline
FT \citep{kim2018paraphrasing} & 66.47\am0.09 & 66.05\am0.40 & 69.55\am0.33 & 67.28\am0.17 & 68.05\am0.42 & 69.86\am0.24 \\
FT + \ours & 66.58\am0.12 & 66.55\am0.39 & 69.85\am0.23 & 67.54\am0.19 & 69.03\am0.33 & 70.91\am0.30 \\
$\Delta$ & 0.11 & \color{orange}{0.50} & \color{orange}{0.30} & \color{orange}{0.26} & \color{orange}{0.98} & \color{orange}{1.05} \\
\hline
LS-KD \citep{sun2024logit} & 63.38\am0.29 & 62.66\am0.29 & 63.49\am0.06 & 66.66\am0.10 & 65.72\am0.10 & 66.58\am0.17 \\
LS-KD + \ours & 67.17\am0.08 & 67.28\am0.18 & \textbf{70.33}\am0.14 & \textbf{68.99}\am0.23 & \textbf{69.80}\am0.18 & \textbf{71.48}\am0.28 \\
$\Delta$ & \color{orange}{3.79} & \color{orange}{4.62} & \color{orange}{6.84} & \color{orange}{2.33} & \color{orange}{4.08} & \color{orange}{4.90} \\
\hline
\hline
Entropy \citep{namdari2019review} & 60.24\am0.23 & 60.29\am0.19 & 62.37\am0.23 & 64.37\am0.9 & 62.47\am0.21 & 63.17\am0.23 \\

MC dropout \citep{gal2016dropout} & 63.59\am0.13 & 63.42\am0.27 & 67.94\am0.05 & 67.90\am0.36 & 69.64\am0.11 & 70.23\am0.24 \\

MSP \citep{pearce2021understanding} & 60.71\am0.19 & 60.62\am0.16 & 62.77\am0.36 & 64.56\am0.23 & 62.80\am0.22 & 63.70\am0.10 \\

PTLoss \citep{zhang2024knowledge} & 65.96\am0.18 & 66.28\am0.16 & 68.27\am0.02 & 66.52\am0.19 & 68.59\am0.21 & 69.35\am0.31 \\

EA-KD \citep{su2025eakdentropybasedadaptiveknowledge} & 66.30\am0.20 & 66.52\am0.30 & 69.04\am0.15 & 67.05\am0.07 & 69.37\am0.23 & 70.33\am0.20 \\
\hline
\end{tabular}
}
\label{tab:under}%
\vspace{-10pt}
\end{table*}%

\vspace{-2pt}
\subparagraph{Experimental Results.}
We present the results in Table \ref{tab:under}. Following the protocol outlined in \citep{sun2024logit}, we highlight in orange the improvements greater than 0.15, indicating non-trivial enhancements. Traditional knowledge distillation methods typically assume that the teacher model is reliable and superior to the student. However, as shown in Table \ref{tab:under}, when the teacher performs worse than the student under domain shift, following the teacher will result in students that perform worse than those trained from scratch. In contrast, integrating \ours~not only improves model performance by up to $10.89$\% but also enables the student to surpass that from scratch. These findings highlight \ours’s effectiveness in enhancing supervised learning, leveraging a pretrained model as guidance, even when the guidance is unreliable. By selectively leveraging useful ``dark knowledge'' while avoiding misleading supervision, \ours~ensures robust model learning.

We also compare against other uncertainty-aware KD methods \citep{su2025eakdentropybasedadaptiveknowledge, zhang2024knowledge, pearce2021understanding, namdari2019review, gal2016dropout}. Results show that KD combined with \ours~outperforms these baselines. This is because existing methods primarily rely on heuristic uncertainty estimates, which can be overconfident and poorly calibrated, particularly under domain shifts. In contrast, \ours~can provide more adaptive and reliable guidance. Moreover, methods such as MC dropout require multiple forward passes, which is computationally expensive. Please refer to Appendix \ref{app: comp} for a comparison of the computation overhead between \ours~and MC dropout.

\vspace{-5pt}
\subparagraph{Ablation Studies.} We evaluate the performance of a heterogeneous teacher-student framework and present the results in Table \ref{tab:hete} in Appendix \ref{app:hete}. The results show that, for all knowledge distillation methods, performance improves when combined with \ours, further validating the effectiveness of our approach across different teacher-student structures. 

We also explore another hard version of the weighting function: $w=1$ if $u=0$, $w=0$ if $u>0$ \citep{vovk2016criteria}, which is similarly effective, please see the results in Appendix \ref{app: hard_w}. The rationale for the hard weighting function follows \citep{vovk2016criteria}, which aims to ensure that prediction sets are as close as possible to single-element sets, making them more informative.

\vspace{-5pt}

\paragraph{Semi-Supervised Image Classification.}
We evaluate the effectiveness of our framework in improving the performance of semi-supervised learning methods, regarding classification tasks.
\vspace{-5pt}
\subparagraph{Experimental Setup.}
We conduct experiments on several SSL image classification benchmarks, including CIFAR-10/100 \citep{krizhevsky2009learning} and STL-10 \citep{coates2011analysis}. For all experiments, we report the mean and standard deviation over four repeated runs. We measure the Top‑1 accuracy for a range of baselines, including UDA \citep{xie2020unsupervised}, FixMatch \citep{sohn2020fixmatch} and FlexMatch \citep{zhang2021flexmatch}, evaluating each both with and without \ours. For an unlabeled image $x$, we construct the prediction set $\mathcal{C}(x)$ for pseudo-labeling using confidence score as described in Section \ref{sec:pre} with $\alpha=0.05$. The associated uncertainty is defined as $u(x) = \frac{|\mathcal{C}(x)|-1}{K-1}$ \citep{vovk2016criteria}, where $K$ is the total number of classes. The adaptive weight is given by $w=\exp(-\gamma u)$ with $\gamma=8.0$. Please refer to Appendix \ref{hyper_ssl} for a sensitivity analysis of the hyperparameter choices and Appendix \ref{app:ssl_train} for the training details. 

\begin{table*}[ht]
\caption{Top-1 accuracy (\%) of various baselines with and without \ours~on several semi-supervised image classification benchmarks, using cross-entropy as the guidance loss. We use $\Delta$ to show mean performance gain relative to conventional methods without \ours, observing upto {\bf +5.98\%} higher accuracy.}
\centering
\resizebox{\textwidth}{!}{
\begin{tabular}{lccccccccc}
\hline
\multirow{2}{*}{Approach} & \multicolumn{3}{c}{CIFAR-10} & \multicolumn{3}{c}{CIFAR-100} & \multicolumn{3}{c}{STL-10} \\ \cline{2-10} 
 & 40 labels & 250 labels & 4000 labels & 400 labels & 2500 labels & 10000 labels & 40 labels & 250 labels & 1000 labels \\ \hline
UDA \citep{xie2020unsupervised}  & 57.73\am6.98 & 89.44\am1.57 & 91.86\am0.75 & 25.95\am1.07 & 57.57\am0.12 & 66.94\am0.16 & 53.88\am0.58 & 76.91\am0.10 & 87.59\am0.34 \\ 
UDA + \ours & 61.28\am6.33 & 92.69\am0.12 & 93.17\am0.10 & 28.02\am0.54 & 58.54\am0.80 & 67.50\am0.35 & 54.70\am0.51 & 77.45\am0.12 & 88.31\am0.10 \\
$\Delta$   & 3.55 & 3.25 & 1.31 & 3.07 & 0.97 & 0.56 & 0.82 & 0.54 & 0.72 \\ 
\hline
FixMatch \citep{sohn2020fixmatch} & 64.18\am4.57  & 89.97\am1.04  & 91.29\am0.65  & 40.36\am0.83 & 61.14\am0.40  & 67.50\am0.85  & 58.03\am1.28  & 78.89\am0.46  & 88.54\am0.10  
\\ 
FixMatch + \ours   & 70.16\am3.34  & 92.23\am0.72  & 93.97\am0.11  & 41.98\am0.55  & 63.41\am1.46  & 70.03\am0.29  & 62.70\am0.84  & 80.83\am0.39  & 89.35\am0.10  \\
$\Delta$   & 5.98 & 2.26 & 2.66 & 1.62 & 2.27 & 2.43 & 4.67 & 1.94 & 0.81 \\ 
\hline
FlexMatch \citep{zhang2021flexmatch} & 73.24\am1.61  & 90.62\am0.49  & 92.11\am0.47  & 51.25\am1.63  & 63.59\am1.03  & 71.63\am0.48  & 62.55\am2.22  & 82.63\am1.20  & 89.94\am0.20  \\ 
FlexMatch + \ours & 76.98\am0.45  & 92.89\am0.10  & 94.28\am0.10  & 55.63\am1.20  & 69.22\am0.52  & 72.71\am0.28  & 65.98\am1.55  & 83.94\am0.23  & 92.27\am0.10  \\ 
$\Delta$   & 3.74 & 2.27 & 2.17 & 4.38 & 5.63 & 1.08 & 3.43 & 1.31 & 2.33 \\ 
\hline
\end{tabular}
}
\label{tab:ssl}
\vspace{-10pt}
\end{table*}

\begin{wraptable}{r}{0.5\textwidth}
\vspace{-13pt}
\centering
\caption{Top-1 accuracy (\%) on CIFAR-100 for semi-supervised image classification, comparing different baselines with and without \ours, using MSE as the guidance loss. $\Delta$ denotes the mean performance gain over the corresponding baseline without \ours.}
\resizebox{\linewidth}{!}{
\begin{tabular}{lccc}
\toprule
Approach & 400 labels & 2500 labels & 10000 labels \\
\midrule
UDA \citep{xie2020unsupervised} & $6.36 \pm 0.47$ & $31.48 \pm 0.38$ & $57.50 \pm 0.31$ \\
UDA+\ours & $7.76 \pm 0.15$ & $32.60 \pm 0.21$ & $59.79 \pm 0.11$ \\
$\Delta$ & $1.40$ & $1.12$ & $2.29$ \\
\midrule
FixMatch \citep{sohn2020fixmatch} & $8.56 \pm 0.26$ & $33.14 \pm 0.69$ & $60.93 \pm 0.41$ \\
FixMatch+\ours & $9.82 \pm 0.62$ & $35.36 \pm 1.03$ & $61.74 \pm 0.20$ \\
$\Delta$ & $1.26$ & $2.22$ & $0.81$ \\
\midrule
FlexMatch \citep{zhang2021flexmatch} & $10.04 \pm 0.24$ & $36.10 \pm 0.36$ & $61.36 \pm 0.07$ \\
FlexMatch+\ours & $11.07 \pm 0.55$ & $38.87 \pm 0.82$ & $62.80 \pm 0.21$ \\
$\Delta$ & $1.03$ & $2.77$ & $1.44$ \\
\bottomrule
\end{tabular}
}
\vspace{-12pt}
 \label{tab:mse}
\end{wraptable}

\subparagraph{Experimental Results.}
We present the results in Table~\ref{tab:ssl}. As shown, integrating \ours~consistently improves performance across all baselines. This highlights the effectiveness of \ours~in semi-supervised learning. By adaptively reweighting the influence of pseudo-labels, \ours~reduces reliance on noisy supervision, mitigating error propagation and leading to improved overall performance.

As ablation studies, in addition to the commonly used cross-entropy loss for guidance in SSL, we investigate the mean-squared error (MSE) loss as an alternative. Specifically, we apply the MSE loss between the logits of a strongly augmented input $x_\text{strong}$ and its weakly augmented counterpart $x_\text{weak}$. As shown in Table~\ref{tab:mse}, our method remains effective under this alternative formulation and continues to yield performance improvements.

% \begin{table}[ht]
% \centering
% \caption{Top-1 accuracy (\%) on CIFAR-100 for semi-supervised image classification, comparing various baselines with and without \ours~using MSE as the guidance loss. $\Delta$ denotes the mean performance gain over the corresponding baseline without \ours.}
% \begin{tabular}{lccc}
% \toprule
% Approach & 400 labels & 2500 labels & 10000 labels \\
% \midrule
% UDA & $6.36 \pm 0.47$ & $31.48 \pm 0.38$ & $57.50 \pm 0.31$ \\
% UDA+AdaConG & $7.76 \pm 0.15$ & $32.60 \pm 0.21$ & $59.79 \pm 0.11$ \\
% $\Delta$ & $1.40$ & $1.12$ & $2.29$ \\
% \midrule
% FixMatch & $8.56 \pm 0.26$ & $33.14 \pm 0.69$ & $60.93 \pm 0.41$ \\
% FixMatch+AdaConG & $9.82 \pm 0.62$ & $35.36 \pm 1.03$ & $61.74 \pm 0.20$ \\
% $\Delta$ & $1.26$ & $2.22$ & $0.81$ \\
% \midrule
% FlexMatch & $10.04 \pm 0.24$ & $36.10 \pm 0.36$ & $61.36 \pm 0.07$ \\
% FlexMatch+AdaConG & $11.07 \pm 0.55$ & $38.87 \pm 0.82$ & $62.80 \pm 0.21$ \\
% $\Delta$ & $1.03$ & $2.77$ & $1.44$ \\
% \bottomrule
% \end{tabular}
% \vspace{-10pt}
% \end{table}

% \vspace{-5pt}
\paragraph{Gridworld Navigation.}
\vspace{-5pt}
We investigate the use of reinforcement learning for solving gridworld navigation tasks, leveraging a pretrained imitation policy as prior guidance. We demonstrate how \ours~improves policy learning efficiency and robustness in challenging and unseen environments when the imitation policy is limited due to generalization constraints.

\vspace{-8pt}
\subparagraph{Experimental Setup.}
We evaluate \ours~across three gridworld environment \citep{chevalier2024minigrid} scenarios, as illustrated in Fig. \ref{fig:rl_env}. For the environment details, please refer to Appendix \ref{app:rl}. We collect expert demonstration for the Lava 1 and Door environments to train IL policies via behavior cloning. After training the IL policy $\pi_\text{I}$, we utilize it to guide the training of the RL policy $\pi_\text{R}$. The Lava 2 environment represents a shifted variant of Lava 1, featuring modified environmental configurations. Importantly, we do not collect expert demonstration for Lava 2, and no IL policy is trained on this environment. For a given state $s$, the guidance weight is defined as: $w(s)=\frac{\exp(-u_\text{I}(s))}{\exp(-u_\text{I}(s)) + \exp(-u_\text{R}(s))}$, where $u_\text{I}$ and $u_\text{R}$ are the prediction uncertainties of the IL and RL policies as described in Section \ref{sec:rl}. And we sample the action $a\in\{a_\text{I}, a_\text{R}\}$ to take according to the distribution induced by this guidance weight. Furthermore, we explore another hard variant of \ours, instead of defining $w(s)$ as a probability distribution informed by the relative uncertainties of the IL and RL policies, we take \text{argmax} to compare IL and RL prediction uncertainties: $w(s)=1$ when $u_\text{I}(s) < u_\text{R}(s)$, otherwise $w(s)=0$. Based on $w(s)$, we dynamically decide which action to take: $a = a_\text{I}$ if $w(s)=1$, otherwise $a = a_\text{R}$.

\vspace{-6pt}
\subparagraph{Baselines.}
We compare \ours~and Hard \ours~against several baselines, including (1) \textbf{Soft Actor Critic (SAC)} \citep{haarnoja2018soft}, a purely RL approach, (2) \textbf{IBRL} \citep{hu2023imitation}, which leverages a pretrained imitation learning (IL) model to bootstrap RL. During the training process, IBRL queries the target Q-network and selects actions by comparing the Q-values of two candidate actions and taking the action with the higher Q-value, and (3) \textbf{Soft IBRL} \citep{hu2023imitation}, a probabilistic variant of IBRL. Instead of selecting the action via a hard \text{argmax}, Soft IBRL samples the action according to a distribution proportional to the Q-values. 

\begin{figure}[t]
\vspace{-5pt}
     \centering
     \begin{subfigure}[b]{0.245\linewidth}
         \centering
         \includegraphics[width=\textwidth]{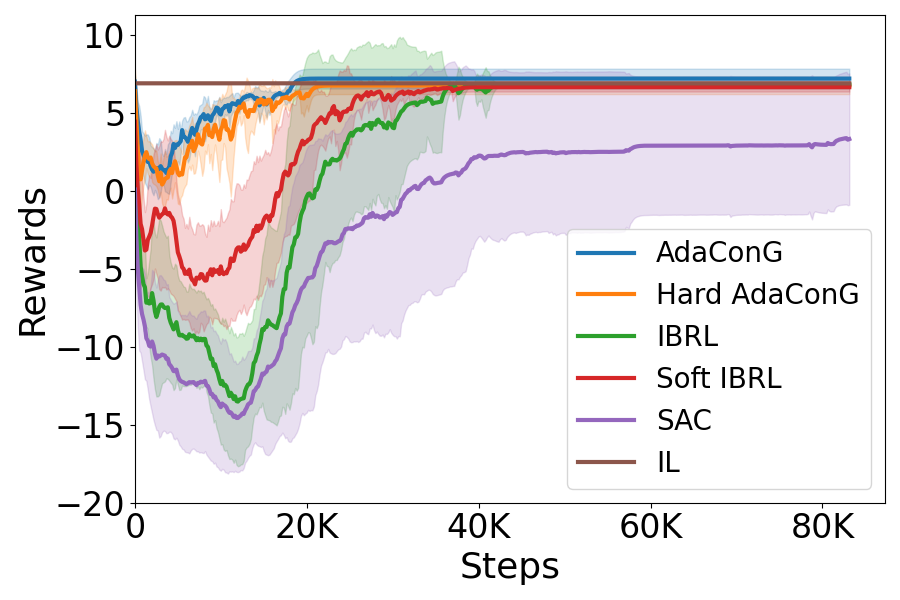}
         \caption{Lava 1}
         \label{fig:reward_lava1}
     \end{subfigure}
     \hfill
     \begin{subfigure}[b]{0.245\linewidth}
         \centering
         \includegraphics[width=\textwidth]{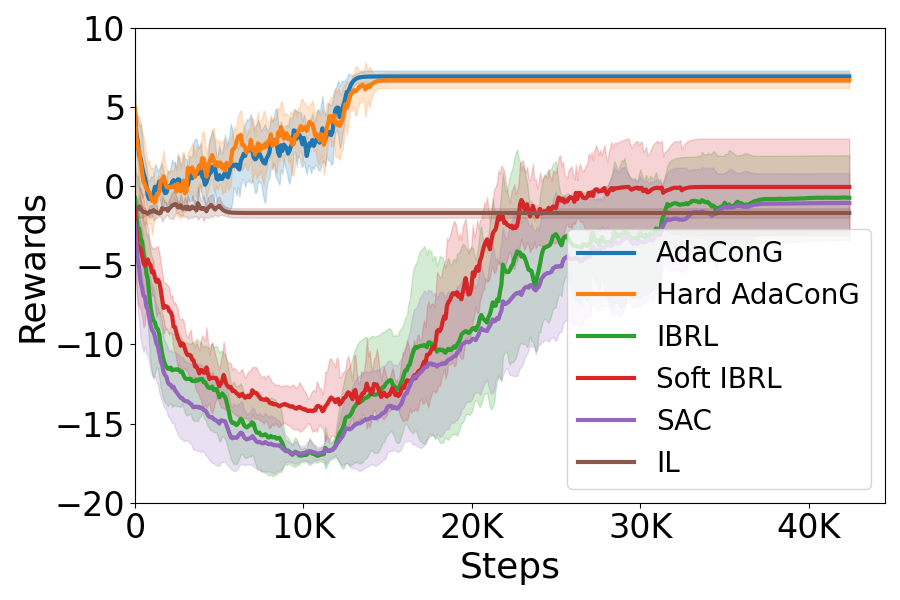}
         \caption{Lava 2}
         \label{fig:reward_lava2}
     \end{subfigure}
     \hfill
     \begin{subfigure}[b]{0.245\linewidth}
         \centering
         \includegraphics[width=\textwidth]{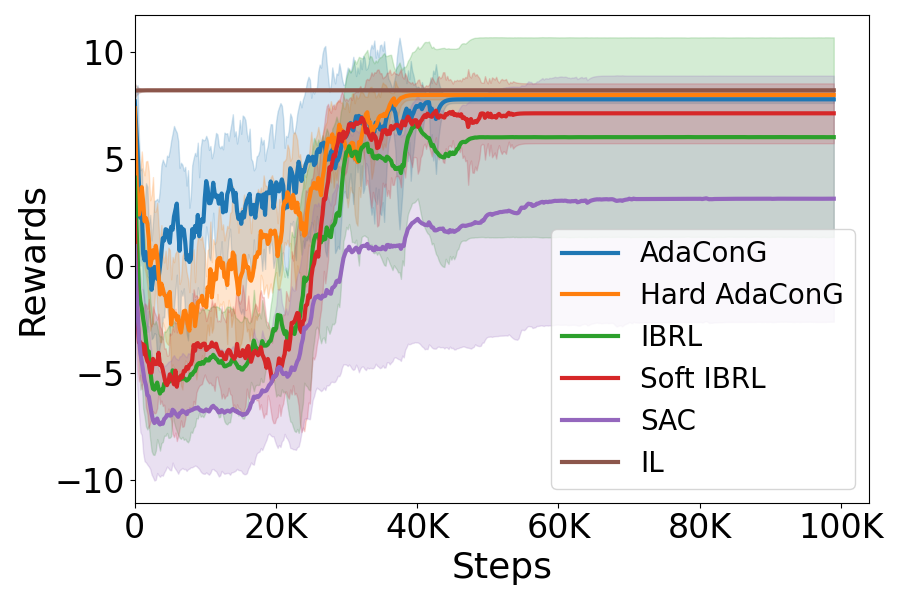}
         \caption{Door}
         \label{fig:reward_door}
     \end{subfigure}
     \hfill
     \begin{subfigure}[b]{0.245\linewidth}
         \centering
         \includegraphics[width=\textwidth]{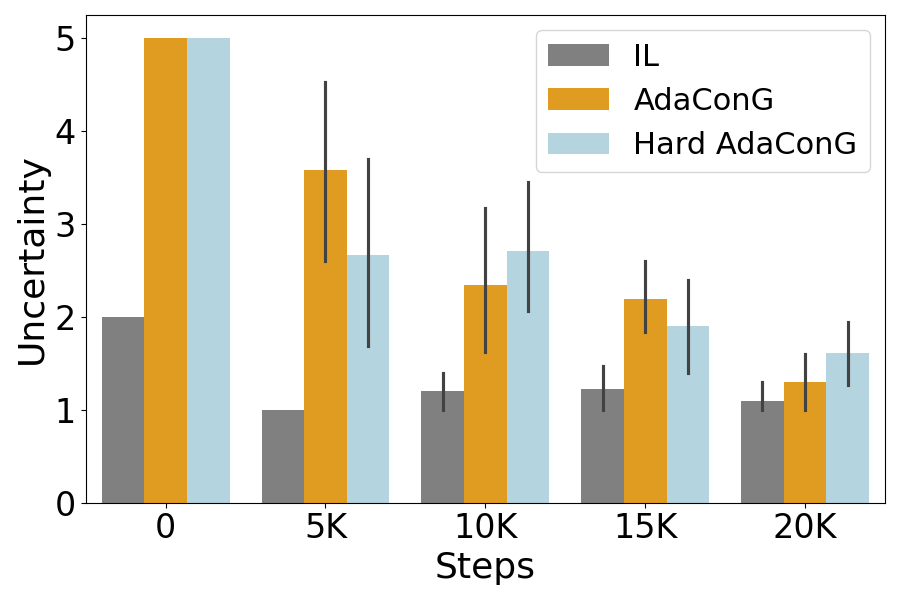}
         \caption{Prediction Uncertainty}
         \label{fig:rl_unc}
     \end{subfigure}
     \vspace{-10pt}
        \caption{\textbf{(a-c) Learning Curves.} We compare \ours~and Hard \ours~with other baselines, including SAC, IBRL, and Soft IBRL, and present their learning curves across three environments: (a) Lava 1, (b) Lava 2, and (c) Door. \ours~and Hard \ours~perform similarly, converging faster and achieving higher rewards than other baselines in all environments. \textbf{(d) Prediction Uncertainty.} We show the average prediction uncertainties of \ours~and Hard~\ours, taking the Lava 1 environment as the example. Over time, their prediction uncertainties shrink and approach that of the IL policy, demonstrating the development of a well-learned RL policy.}
        \label{fig:rl_results}
        \vspace{-15pt}
\end{figure}

\vspace{-6pt}
\subparagraph{Experimental Results.}
We run all experiments across ten random seeds and present the results in Fig. \ref{fig:rl_results}. First, we compare the learning curves of \ours~and Hard \ours~against other baselines across three environments: Lava 1, Lava 2, and Door. Both \ours~and Hard \ours~demonstrate similar performance, converging faster and achieving higher rewards than all other baselines. Before the agent reaches the goal, the reward function is defined as the negative Manhattan distance between the agent's current location and the goal, normalized by the maximum step limit of 100. Consequently, the accumulated episode rewards initially decrease as the agent explores the environment and accrues negative rewards but increase as it learns. The rewards of \ours~and Hard \ours~are consistently higher than those of other baselines while converging faster. This can be attributed to the efficiency of \ours, as it compares the prediction uncertainties of teacher and student models instead of relying on Q-values. Methods like IBRL and Soft IBRL, which depend on Q-values to decide between IL or RL actions, may make suboptimal decisions initially due to poorly trained Q-networks. For instance, even if an IL action is superior, its Q-value might be lower than that of an RL action.

In the shifted environment Lava 2, the overall rewards of the IL policy are lower due to generalization constraints. IBRL and Soft IBRL rewards eventually converge close to the IL policy’s performance, as these methods are not uncertainty-aware. Blindly relying on a IL policy underperforming due to environment shifts can lead to suboptimal performance. In contrast, \ours~and Hard \ours~consider the IL policy's prediction uncertainty, enabling faster convergence and achieving rewards over $6\times$ higher than the best-performing baselines after convergence. When the IL policy’s predictions are confident, the RL agent relies more on them; otherwise, the RL agent explores independently. Even though the IL policy's overall reward is not high, it still provides useful knowledge. This allows the RL agent to learn from the IL policy and eventually achieve higher rewards than the IL policy. 

We also show the average policy prediction uncertainties of \ours~and Hard \ours, for Lava 1 environment, in Fig. \ref{fig:rl_unc}. Over time, their prediction uncertainties decrease and approach that of the IL policy, demonstrating the progression toward a well-learned RL policy.

\vspace{-6pt}
\paragraph{Autonomous Driving.}
This task involves learning a steering prediction policy in autonomous driving for an RGB-only input model, guided by a pretrained multi-modal teacher to transfer knowledge to the student mdoel. We evaluate the effectiveness of different knowledge distillation methods under domain shifts and sensor noise, comparing performance with and without the use of \ours.

\vspace{-5pt}
\begin{wraptable}{r}{0.5\textwidth}
\vspace{-13pt}
\caption{Mean accuracy (\%) of steer prediction of different knowledge transfer methods with and without~\ours~under domain shifts.}
\centering
\resizebox{\linewidth}{!}{
\begin{tabular}{lccc}
\hline
\multirow{2}{*}{Approach} & \multicolumn{3}{c}{Mean Accuracy (\%)} \\ \cline{2-4} 
 & \multicolumn{1}{l}{without \ours} & \multicolumn{1}{l}{with \ours} & \multicolumn{1}{l}{$\Delta$} \\ 
\hline
KD  & 73.5 & 76.8 & 3.3\\
FitNet  & 72.4 & 76.2 & 3.8 \\
PKT  & 72.8 & 75.9 & 3.1 \\
FT  & 73.1 & 76.4 & 3.3 \\

% \hline
% EA-KD \cite{romero2014fitnets} & 72.4 &  &  \\
% PTLoss \cite{passalis2018learning} & 72.8 &  &  \\
\hline
Teacher (RGB+Depth+Edge) & 78.5 & -- & -- \\
Student (RGB) & 71.8 & -- & -- \\
\hline
\end{tabular}
}
\label{tab:steer}%
\vspace{-5pt}
\end{wraptable}%

\subparagraph{Experimental Setup.} 
We adopt mean accuracy (mAcc) as the evaluation metric for the task of steer prediction, following prior works \citep{shen2021gradient, shen2023auxiliary, shen2024task}. We use the real-world driving dataset SullyChen \citep{chen2018collection} for evaluation, which includes diverse driving scenarios with various road types and conditions. We use Nvidia PilotNet \citep{bojarski2016end} as the backbone for both the teacher and student models. The teacher model is a multi-modal network that takes RGB images, depth, and edge maps as input, while the student model is unimodal, relying solely on RGB images. For more details of the setup, please see Appendix \ref{app:ad_set}. We first train the teacher model $f_p$ offline. Then we use it to guide the student model $f_t$ learning, while the RGB images for $f_t$ training have domain shifts by Gaussian noise corruption compared to the ones used for $f_p$ training. Detailed information about domain shifts and model training can be found in Appendices \ref{app:ad_data} and \ref{app:ad_train}, respectively. We evaluate multiple knowledge distillation methods as baselines, comparing their performance with and without the integration of \ours, including KD \citep{hinton2015distilling}, FitNet \citep{romero2014fitnets}, PKT \citep{passalis2018learning}, and FT \citep{kim2018paraphrasing}. 

\vspace{-5pt}
\subparagraph{Experimental Results.}
We report the mean accuracy of steer prediction for various KD methods under domain shifts with and without \ours~in Table \ref{tab:steer}. As observed, incorporating \ours~consistently enhances the accuracy. This demonstrates the effectiveness of \ours~in improving model performance, as its adaptive guidance mechanism strategically prevents over-reliance on uncertain teacher predictions, facilitating more reliable and robust target model learning.
    
\vspace{-8pt}
\section{Conclusions}
\vspace{-5pt}
We propose \ours, an approach for learning with guidance under uncertainty. \ours~integrates split conformal prediction to adaptively modulate the influence of guidance signals based on their associated uncertainty. By selectively leveraging reliable signals and filtering out misleading supervision, \ours~enables effective learning even in the presence of noise. Unlike conventional methods that assume guidance is always trustworthy, \ours~can still extract useful “dark knowledge” under uncertainty. The framework is simple yet effective, and broadly applicable to a wide range of tasks. We validate \ours~across diverse settings and tasks including knowledge distillation, semi-supervised image classification, gridworld navigation, and autonomous driving, demonstrating improved performance and robustness. For a discussion on future work, please see Appendix \ref{app:limit}.

% \section{Reproducibility Statement}
% To ensure reproducibility, we provide detailed descriptions of all training setups in the experiment Section~\ref{sec:exp}, with additional specifications in Appendix~\ref{app:img}, Appendix~\ref{app:ssl_train}, Appendix~\ref{app:rl}, and Appendix~\ref{app:ad_set}. All datasets used are publicly available, and we will release the code upon publication.

\bibliography{reference}
\bibliographystyle{plainnat}

%%%%%%%%%%%%%%%%%%%%%%%%%%%%%%%%%%%%%%%%%%%%%%%%%%%%%%%%%%%%%%%%%%%%%%%%%%%%%%%
%%%%%%%%%%%%%%%%%%%%%%%%%%%%%%%%%%%%%%%%%%%%%%%%%%%%%%%%%%%%%%%%%%%%%%%%%%%%%%%
% APPENDIX
%%%%%%%%%%%%%%%%%%%%%%%%%%%%%%%%%%%%%%%%%%%%%%%%%%%%%%%%%%%%%%%%%%%%%%%%%%%%%%%
%%%%%%%%%%%%%%%%%%%%%%%%%%%%%%%%%%%%%%%%%%%%%%%%%%%%%%%%%%%%%%%%%%%%%%%%%%%%%%%
\newpage
\appendix
\onecolumn

\section{Appendix}

\subsection{Knowledge Distillation} \label{app:img}
We evaluate two settings: (1) Homogeneous Structure, where both the teacher and student models share the same type of architecture (e.g., ResNet-32x4 and ResNet-8x4), and (2) Heterogeneous Structure, where the teacher and student models are of different architectures (e.g., ResNet-32x4 and ShuffleNet-V1). We evaluate a wide range of neural network architectures, including ResNet \citep{he2016deep}, WRN \citep{zagoruyko2016wide}, VGG \citep{simonyan2014very}, ShuffleNet-V1 \citep{zhang2018shufflenet}/V2 \citep{ma2018shufflenet}, and MobileNet-V2 \citep{sandler2018mobilenetv2}.
 
\subsubsection{Dataset Details} \label{app:img_noise}
We conduct our experiments on the CIFAR-100 dataset \cite{krizhevsky2009learning}, which consists of 60K images, 50K for training and 10K for testing, across 100 distinct categories. We introduce domain shifts to the dataset for training the target model. We add Gaussian noise with zero mean and a standard deviation of 0.05 to $40\%$ of the training data of 50K images, where the noisy samples are selected uniformly at random across the entire dataset to ensure consistent noise distribution. Then we shuffle the dataset and randomly split it into a $90\%$ training set $\mathcal{D}_\text{train}^t$, and a $10\%$ calibration set $D_\text{cal}$, ensuring that both sets are drawn from the same underlying distribution. Additionally, the same Gaussian noise is added to $40\%$ of the testing data of 10K images, to form the noisy test set $D_\text{test}$, which allows us to evaluate the performance of the target model. 

\subsubsection{Heterogeneous Teacher-Student Structure} \label{app:hete}
\vspace{-5pt}
\begin{table*}[htb]
\caption{Top-1 accuracy (\%) of various knowledge distillation methods with and without \ours~on CIFAR-100 under heterogeneous structure. We use $\Delta$ to show mean performance gain relative to conventional knowledge distillation methods without \ours. We highlight in orange deltas greater than 0.15, indicating non-trivial enhancement following the protocol in \cite{sun2024logit}.}
\centering
\resizebox{0.7\textwidth}{!}{
\begin{tabular}{lccc}
\hline \multirow{2}{*}{Teacher} & ResNet50 & VGG13 & WRN-40-2 \\
& 62.79  & 61.47 & 58.76 \\
Student & ShuffleNet-V1 & MobileNet-V2 & ShuffleNet-V2 \\
& 64.52\am0.62 & 56.47\am0.03 & 66.35\am0.12 \\
\hline 
KD \cite{hinton2015distilling} & 58.30\am0.24  & 53.08\am0.57 & 59.61\am0.03 \\
KD + \ours & 65.71\am0.36 & 57.69\am0.52 & 67.57\am0.22 \\
$\Delta$ & \color{orange}{7.41} & \color{orange}{4.41} & \color{orange}{7.96}\\
\hline 
FitNet \cite{romero2014fitnets} & 63.97\am0.25 & 54.77\am0.40 & 66.03\am0.48 \\
FitNet + \ours & 64.49\am0.15 & 55.75\am0.37 & 67.69\am0.11 \\
$\Delta$ & \color{orange}{0.52} & \color{orange}{0.98} & \color{orange}{1.66}\\
\hline
PKT \cite{passalis2018learning} & 66.26\am0.26 & 56.53\am0.13 & 66.38\am0.18 \\
PKT + \ours & 66.65\am0.16 & 57.01\am0.18 & 67.88\am0.42 \\
$\Delta$ & \color{orange}{0.39} & \color{orange}{0.48} & \color{orange}{1.50} \\
\hline
FT \cite{kim2018paraphrasing} & 63.85\am0.34 & 56.36\am0.49 & 66.34\am0.21 \\
FT + \ours & 65.13\am0.43 & 57.40\am0.20 & 67.29\am0.12 \\
$\Delta$ & \color{orange}{1.28} & \color{orange}{1.04} & \color{orange}{0.95}\\
\hline
\end{tabular}
}
\label{tab:hete}%
\vspace{-5pt}
\end{table*}%

As part of our ablation studies, we evaluate the performance of a heterogeneous teacher-student framework and present the results in the following Table \ref{tab:hete}. The table shows that, for all knowledge transfer methods, performance improves when combined with \ours, further validating the effectiveness of our approach across different teacher-student structures.

\subsubsection{Hyperparameter Sensitivity Analysis} \label{hyper_sp}
We conduct sensitivity analysis for the design choices of the key hyperparameters, including the error rate $\alpha$ for split CP and the temperature $\gamma$ of the adaptive weighting $w=\exp(-\gamma u)$. We use ResNet-110 as the pretrained teacher and ResNet-20 as the student model to train. 

\begin{figure}[t]
    \centering
    \includegraphics[width=\linewidth]{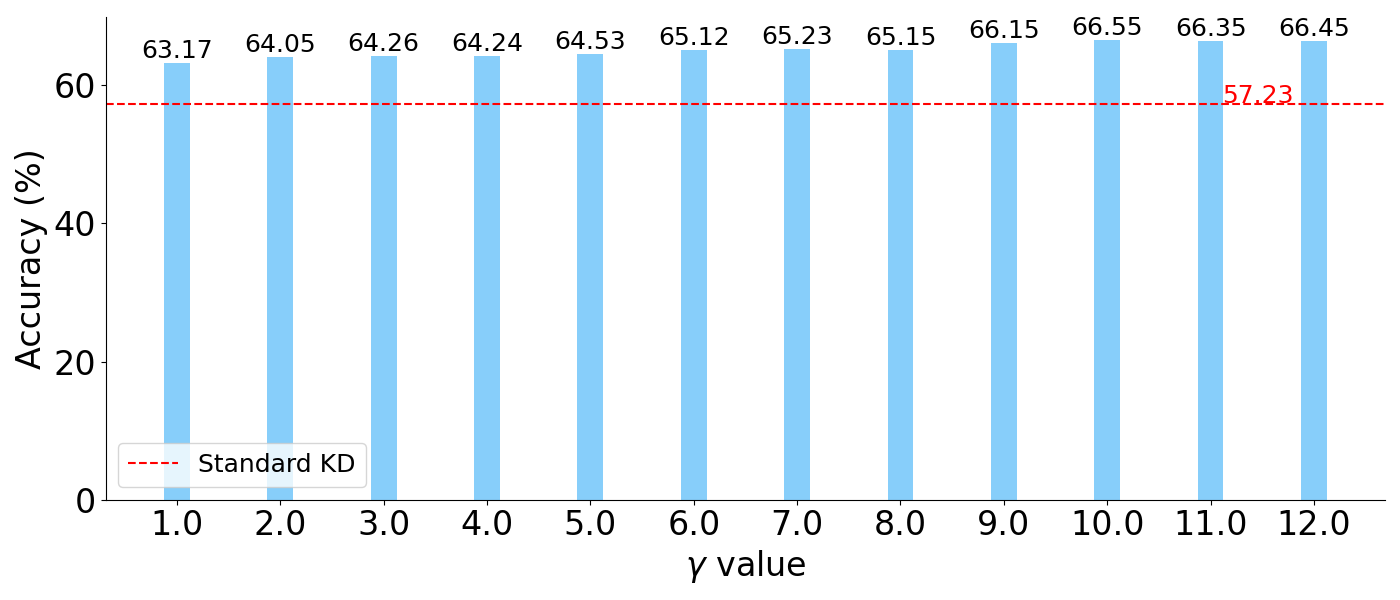}
    \caption{Top-1 accuracy of KD using \ours~with varying temperature $\gamma$ values for adaptive weighting.}
    \label{fig:kd_g}
\end{figure}

We first analyze the sensitivity of the temperature parameter $\gamma$ in the adaptive weighting function. The results are presented in Fig. \ref{fig:kd_g}, showing how $\gamma$ affects the student’s Top-1 accuracy with $\alpha = 0.1$. Across all tested values, combining KD with \ours~consistently outperforms standard KD. We observe that accuracy generally increases as $\gamma$ increases. This trend is intuitive, when $\gamma$ is too small, the exponential decay used for reweighting may not sufficiently suppress the influence of noisy teacher predictions. We select $\gamma = 10.0$ as our default setting, since performance tends to not increase for $\gamma > 10.0$.

\begin{figure}[t]
    \centering
    \includegraphics[width=0.5\linewidth]{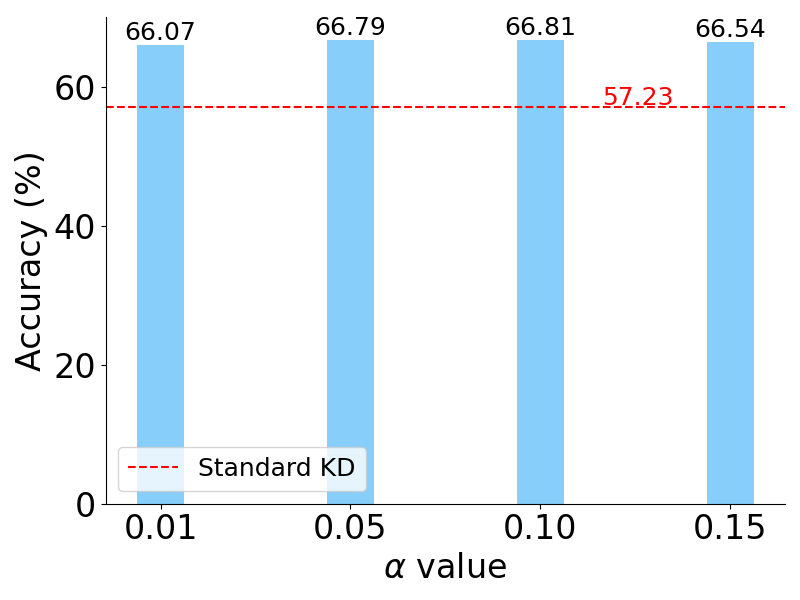}
    \caption{Top-1 accuracy of KD using \ours~with varying $\alpha$ values. The results demonstrate that our approach is robust to the choice of $\alpha$ and consistently outperforms standard KD.}
    \label{fig:kd_a}
\end{figure}

\begin{figure}
    \centering
    \includegraphics[width=0.5\linewidth]{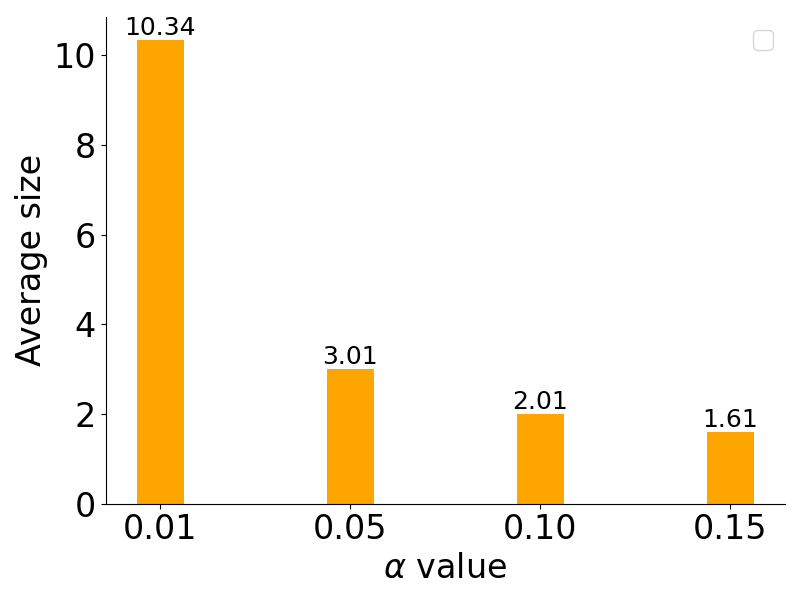}
    \caption{Average size of the prediction set for the teacher model.}
    \label{fig:kd_size}
\end{figure}

We then analyze the sensitivity of $\alpha$. We present the results in Fig. \ref{fig:kd_a}, showing how $\alpha$ influences the student’s Top-1 accuracy with $\gamma=10.0$. The results demonstrate that our approach is robust to the choice of $\alpha$ and consistently outperforms standard KD. We choose $\alpha=0.1$ which yields slightly better results. The underlying insight is as follows: when $\alpha$ is small, the prediction set becomes large, indicating lower teacher confidence. As a result, the teacher's guidance becomes less informative, and the student relies more on its own learning, reducing the benefit of distillation, even when the teacher is accurate. Conversely, when $\alpha$ is large, the prediction set shrinks, making the teacher appear overly confident. In this case, the student may rely too heavily on potentially noisy teacher predictions, leading to suboptimal knowledge transfer. We further support this interpretation by presenting the average size of the prediction set for the teacher model in Fig. \ref{fig:kd_size}, which aligns with our observations.

% \begin{table}[ht]
% \caption{Sensitivity analysis of the error rate $\alpha$ for split CP.}
% \centering
% % \resizebox{\linewidth}{!}{
% \begin{tabular}{lcc}
% \hline
% $\alpha$ & Average $|\mathcal{C}(x)|$ & Accuracy (\%) \\ \hline
% 0.01 & 10.34 & 67.07 \\
% 0.05 & 3.01 & 66.81 \\
% 0.10 & 2.01 & 66.79 \\
% 0.15 & 1.61 & 66.54 \\
% \hline
% \end{tabular}
% \label{tab:alpha_kd}
% \end{table}

% We then analyze the sensitivity of the temperature $\gamma$ for the weighting function. 

% \begin{table}[ht]
% \caption{Sensitivity analysis of the temperature $\gamma$ for the adaptive weighting.}
% \centering
% % \resizebox{\linewidth}{!}{
% \begin{tabular}{lcc}
% \hline
% $\gamma$ & Accuracy \\ \hline
% 1 & 62.17 \\
% 2 & 63.26 \\
% 3 & 63.05 \\
% 4 & 63.24 \\
% 5 & 63.53 \\
% 6 & 64.12 \\
% 7 & 63.93 \\
% 8 & 64.23 \\
% 9 & 64.15 \\
% 10 & 64.73 \\
% \hline
% \end{tabular}
% \label{tab:alpha_kd}
% \end{table}

\subsubsection{Hard Weighting Function} \label{app: hard_w}
We explore another hard version of weighting function: $w=1$ if $u=0$, $w=0$ if $u>0$ \citep{vovk2016criteria}. We use ResNet-110 as the pretrained teacher and ResNet-20 as the student model to train. The rationale for the hard weighting function follows \citep{vovk2016criteria}, which aims to ensure that prediction sets are as close as possible to single-element sets, making them more informative. This scheme is simple and effective, allowing the student to learn from high-confidence teacher predictions while filtering out potentially misleading guidance. Experimental results in Table \ref{tab:hard_w} show that combining this hard weighting scheme with \ours~outperforms standard knowledge distillation methods.

\begin{table*}[htb]
\caption{Top-1 accuracy (\%) of various knowledge distillation methods without and with \ours~using the hard weighting function. We use $\Delta$ to show performance gain relative to conventional knowledge distillation methods and highlight in orange deltas greater than 0.15, indicating non-trivial enhancement following the protocol in \citep{sun2024logit}.}
\centering
% \resizebox{0.7\textwidth}{!}{
\begin{tabular}{lc}
\hline 
% \multirow{2}{*}{Teacher} & Teacher \\
% & 58.78 \\
Approach & Accuracy (\%) \\
% & 66.50  \\
\hline 
KD \citep{hinton2015distilling} & 57.21 \\
KD + \ours & 66.52 \\
$\Delta$ & \color{orange}{9.31} \\
\hline 
FitNet \citep{romero2014fitnets} & 64.65 \\
FitNet + \ours & 66.88 \\
$\Delta$ & \color{orange}{2.23} \\
\hline
PKT \citep{passalis2018learning} & 66.50  \\
PKT + \ours & 66.83 \\
$\Delta$ & \color{orange}{0.33} \\
\hline
FT \citep{kim2018paraphrasing} & 66.47 \\
FT + \ours & 66.69  \\
$\Delta$ & \color{orange}{0.22} \\
\hline
LS-KD \citep{sun2024logit} & 63.40 \\
LS-KD + \ours & 67.11  \\
$\Delta$ & \color{orange}{3.71} \\
\hline
\end{tabular}

\label{tab:hard_w}%
\vspace{-5pt}
\end{table*}

\subsubsection{Direct Use of Nonconformity Scores}
We conduct additional experiments to compare the performance of directly using nonconformity scores versus applying \ours~with quantile computation, as ablation studies, across different teacher/student setups with KD \citep{hinton2015distilling}. The results are shown in Table \ref{tab:direct}, reporting the mean and standard deviation over four repeated runs. As shown, \ours~outperforms the direct use of nonconformity scores. This indicates that computing the quantile of the nonconformity scores is helpful, it is necessary to construct the prediction set for split CP, therefore transferring non conformalized uncertainty measures into rigorous ones.

\begin{table}[ht]
\caption{Top-1 accuracy (\%) on CIFAR-100 for the KD approach, comparing direct use of nonconformity scores versus AdaConG using quantile computation. \ours~outperforms the direct use of nonconformity scores.}
\centering
\resizebox{\textwidth}{!}{
\begin{tabular}{lccccc}
\toprule
Approach & ResNet110/ResNet20 & ResNet56/ResNet20 & ResNet32x4/ResNet8x4 & VGG13/VGG8 & WRN-40-2/WRN-40-1 \\
\midrule
Nonconformity score & $64.17 \pm 0.41$ & $64.90 \pm 0.28$ & $66.89 \pm 0.15$ & $65.19 \pm 0.34$ & $68.06 \pm 0.13$ \\
AdaConG & $66.53 \pm 0.55$ & $66.98 \pm 0.25$ & $68.45 \pm 0.29$ & $67.53 \pm 0.18$ & $69.31 \pm 0.25$ \\
\bottomrule
\end{tabular}
}
\label{tab:direct}
\end{table}

\subsubsection{Training Details} \label{app:img_train}
For the experiments, we use the stochastic gradient descents (SGD) \citep{sutskever2013importance} as the optimizer with momentum 0.9 and weight decay $5e-4$. The epoch number is 240 and the batch size is 128. The initial learning rate is set to 0.01 for MobileNet \citep{sandler2018mobilenetv2}/ShuffleNet \citep{zhang2018shufflenet} architectures and 0.05 for other architectures. The model is trained on an Nvidia RTX 3090 GPU with AMD Ryzen 9 5900 CPU and 32 GB RAM.

\subsubsection{Computation Overhead} \label{app: comp}
We compare the training cost of our approach \ours~against standard knowledge distillation (KD) and KD with MC dropout on an RTX 3090 GPU. For MC dropout,we perform ten forward passes and average the outputs. Compared to MC dropout, the training cost of using \ours~is much lower. Compared to standard KD, the training cost of \ours~is close, since using split CP is just a single pass, which adds minimal computation overhead by a latency of $0.08$ms per sample. 

\begin{table}[ht]
\caption{Comparison of computational overhead between standard KD, KD with \ours, and KD with MC dropout.}
\centering
% \resizebox{\linewidth}{!}{
\begin{tabular}{lcc}
\hline
Approach & Time/Epoch (s) \\ \hline
KD & 6.87 \\
\ours & 7.04 \\
MC dropout & 44.90 \\
\hline
\end{tabular}
\label{tab:comp}
\end{table}

\subsection{Semi-Supervised Image Classification}
\subsubsection{Training Details} \label{app:ssl_train}
Following the setup in \cite{sohn2020fixmatch, zhang2021flexmatch}, we use the WRN-28-8 architecture \cite{zagoruyko2016wide} for the CIFAR-10 and CIFAR-100 datasets \cite{krizhevsky2009learning}, and WRN-37-2 \cite{zagoruyko2016wide} for the STL-10 dataset \cite{coates2011analysis}. We adopt stochastic gradient descent (SGD) \cite{sutskever2013importance} as the optimizer with a momentum of 0.9. The weight decay is set to $5 \times 10^{-4}$ for CIFAR-10 and STL-10, and $1 \times 10^{-3}$ for CIFAR-100. Models are trained for 51,200 iterations with a batch size of 64 and an initial learning rate of 0.03. Experiments are conducted on an Nvidia RTX 3090 GPU with AMD Ryzen 9 5900 CPU and 32 GB RAM.

\subsubsection{Hyperparameter Sensitivity Analysis} \label{hyper_ssl}
We conduct sensitivity analysis for the design choices of the key hyperparameters, including $\alpha$ for split CP and the temperature $\gamma$ for the adaptive weighting $w=\exp(-\gamma u)$. We conduct experiments using the approach FixMatch \citep{sohn2020fixmatch} combined with \ours~on the CIFAR-10 dataset with 40 labels. 

We first analyze the sensitivity of the temperature parameter $\gamma$ in the adaptive weighting function. The results are presented in Fig. \ref{fig:kd_g}, showing how $\gamma$ affects the prediction accuracy with $\alpha = 0.1$. Across all tested values, accuracy generally increases as $\gamma$ increases, which is expected, when $\gamma$ is too small, the exponential decay used for reweighting may not sufficiently suppress the influence of noisy pseudo-labels. We select $\gamma = 8.0$, as performance begins to slightly decline when $\gamma > 8.0$.

\begin{figure}[t]
    \centering
    \includegraphics[width=\linewidth]{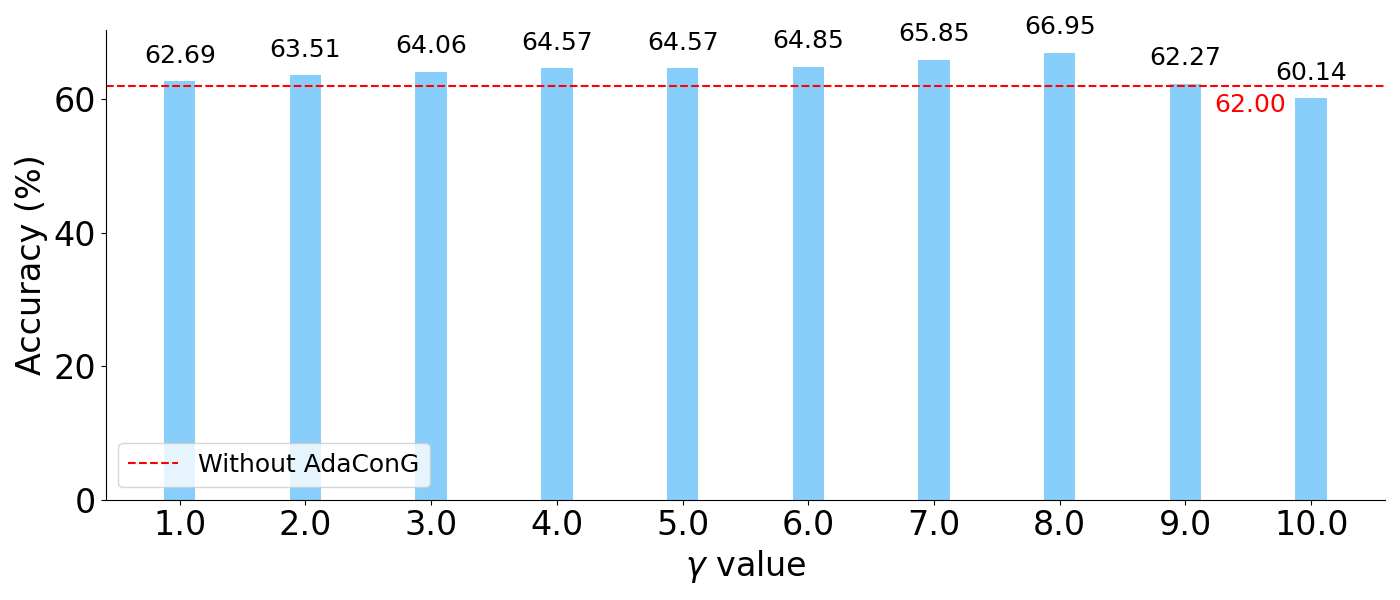}
    \caption{Prediction accuracy of FixMatch using \ours~with varying temperature $\gamma$ values for adaptive weighting.}
    \label{fig:ssl_g}
\end{figure}

\begin{figure}[t]
    \centering
    \includegraphics[width=0.5\linewidth]{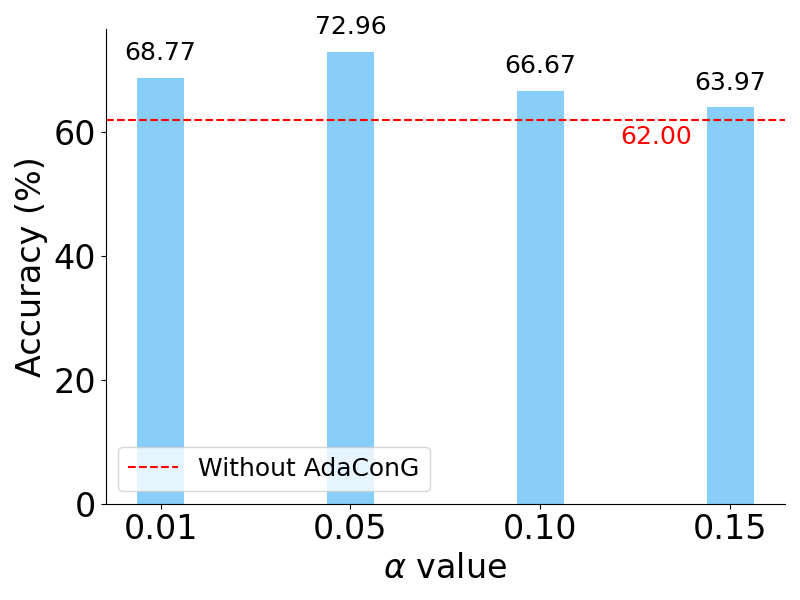}
    \caption{Prediction accuracy of FixMatch using \ours~with varying $\alpha$ values.}
    \label{fig:ssl_a}
\end{figure}

We then analyze the sensitivity of $\alpha$. The results, shown in Fig.\ref{fig:ssl_a}, illustrate how $\alpha$ affects prediction accuracy with $\gamma = 8.0$. FixMatch combined with \ours~outperforms the standard approach across all values. The accuracy first increases and then decreases as $\alpha$ increases, with the highest performance achieved at $\alpha = 0.05$. Therefore, we select $\alpha = 0.05$. The underlying insights are similar to those observed in the knowledge distillation experiments.

% \begin{table}[ht]
% \caption{Sensitivity analysis of the error rate $\alpha$ for split CP.}
% \centering
% % \resizebox{\linewidth}{!}{
% \begin{tabular}{lcc}
% \hline
% $\alpha$ & Average $|\mathcal{C}(x)|$ & Accuracy (\%) \\ \hline
% 0.01 & 10.34 & 64.77 \\
% 0.05 & 3.01 & 73.96 \\
% 0.10 & 2.01 & 59.67 \\
% 0.15 & 1.61 & 52.97 \\
% \hline
% \end{tabular}
% \label{tab:alpha_kd}
% \end{table}

% \begin{table}[ht]
% \caption{Sensitivity analysis of the temperature $\gamma$ for the adaptive weighting.}
% \centering
% % \resizebox{\linewidth}{!}{
% \begin{tabular}{lcc}
% \hline
% $\gamma$ & Accuracy \\ \hline
% 1 & 63.57 \\
% 2 & 63.57 \\
% 3 & 51.69 \\
% 4 & 61.51 \\
% 5 & 63.06 \\
% 6 & 63.85 \\
% 7 & 63.85 \\
% 8 & 65.95 \\
% 9 & 60.27 \\
% 10 & 53.14 \\
% \hline
% \end{tabular}
% \label{tab:alpha_kd}
% \end{table}

\subsection{Imitation-Guided Reinforcement Learning} \label{app:rl}
\begin{figure}[hbt]
\vspace{-5pt}
     \centering
     \begin{subfigure}[bt]{0.22\linewidth}
         \centering
         \includegraphics[width=\linewidth]{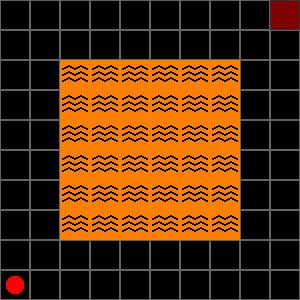}
         \caption{Lava 1}
         \label{fig:env_lava1}
     \end{subfigure}
     \hspace{0.02\textwidth}
     \begin{subfigure}[bt]{0.22\linewidth}
         \centering
         \includegraphics[width=\linewidth]{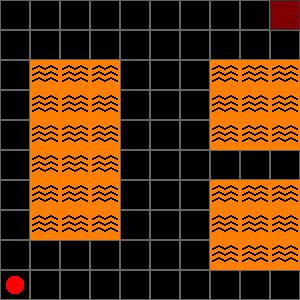}
         \caption{Lava 2}
         \label{fig:env_lava2}
     \end{subfigure}
     \hspace{0.02\textwidth}
     \begin{subfigure}[bt]{0.22\linewidth}
         \centering
         \includegraphics[width=\linewidth]{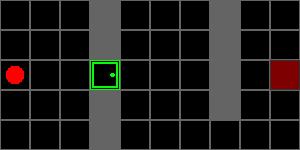}
         \caption{Door}
         \label{fig:env_door1}
     \end{subfigure}
        \caption{\textbf{Gridworld Environment Scenarios.} (a) \textbf{Lava 1}: An autonomous agent (red dot) must navigate to a target position (diagonal square) while avoiding lava regions. (b) \textbf{Lava 2}: A domain-shifted variant of Lava 1 with altered environment dynamics and layout. (c) \textbf{Door}: The agent must traverse a structured environment with doors and walls to reach the designated target position.}
        \label{fig:rl_env}
\end{figure}

The environment scenarios shown in Fig. \ref{fig:rl_env} are adapted from \citep{yu2024beyond} and developed using the Minigrid framework \citep{chevalier2024minigrid}. They are fully observable with discrete state and action spaces. In each environment, the agent's state corresponds to its \text{\{x, y\}} coordinates on the map, and the action space comprises five discrete actions: \text{left}, \text{right}, \text{up}, \text{down}, and \text{stay}. Each episode is capped at a maximum of 100 steps. In the Lava 1 and Lava 2 environments, the reward function is computed as the negative Manhattan distance between the agent's current position and the goal, normalized by the maximum step limit of 100. Upon reaching the goal, the agent receives a terminal reward of \(10-9\times\frac{\text{step count}}{\text{max step}}\). Stepping into the lava results in a reward of -1, and the episode terminates immediately. Similarly, in the Door environment, the reward structure follows the same formulation but without lava, encouraging the agent to minimize its distance to the goal, with the same terminal reward applied upon successful completion.

We collect expert demonstration data comprising state-action pairs for the Lava 1 and Door environments to train imitation learning (IL) models via behavior cloning \citep{torabi2018behavioral}. The demonstration data are inherently uncertain, as multiple valid actions may exist for the same state, as illustrated in \citep{yu2024beyond}, introducing ambiguity in the IL model's predictions.

We calibrate both the IL policy $\pi_\text{I}$ and the RL policy $\pi_\text{R}$, and estimate their prediction uncertainties $u_\text{I}$ and $u_\text{R}$ with $g$ as the identity mapping, $\alpha=0.1$, $N=1000$ and $m=128$, as described in Section \ref{sec:rl}. For a given state $s$, the guidance weight is defined as: $w(s)=\frac{\exp(-u_\text{I}(s))}{\exp(-u_\text{I}(s)) + \exp(-u_\text{R}(s))}$. And we sample the action $a\in\{a_\text{I}, a_\text{R}\}$ to take according to the distribution induced by this guidance weight. To encourage exploration by the RL policy, we define a probability $\epsilon=\min(0.5\frac{t}{S_\text{total}}+0.5\frac{e}{E_\text{total}}, 1)$, where $t$ is the current training step, $e$ is the current episode, $S_\text{total}$ and $E_\text{total}$ are the total training steps and episodes, respectively. At each step, if a random probability $p<\epsilon$, the agent takes an action from the RL policy, otherwise it takes an action based on \ours. As $\epsilon$ increases over time, the agent progressively shifts to a learned RL policy, while initially it relies more on the IL policy through \ours~to facilitate learning. 
 
\subsubsection{Environment Details} \label{app:rl_env}

\subsubsection{Training Details}
For the imitation-guided reinforcement learning experiments, we employ a batch size of 512 and the Adam optimizer \citep{kingma2014adam} with an initial learning rate of $3e{-4}$. For each method and environment scenario, we train for 1000 episodes across 10 different random seeds. The model is trained on an Nvidia RTX 3090 GPU with AMD Ryzen 9 5900 CPU and 32 GB RAM.

% \subsubsection{Hyperparameter Sensitivity Analysis}
% We conduct sensitivity analysis for $\alpha$, evaluating how does it influence the uncertainty estimates of the pretrained IL policy in the target RL environment, by measuring the average prediction uncertainty, which is how many predicted actions for a given state. We present the results in the Fig. below. We observe that the average prediction uncertainty of the IL policy is robust to the choice of $\alpha$, since the action space is not large for the navigation task, which is 5. 

% \begin{figure}[ht]
%     \centering
%     \includegraphics[width=0.5\linewidth]{figs/ad_a.png}
%     \caption{Mean accuracy of KD using \ours~with varying $\alpha$ values.}
%     \label{fig:rl_a}
% \end{figure}

\subsection{Autonomous Driving} 
\subsubsection{Experimental Settings} \label{app:ad_set}
We first define the accuracy with respect to a specific degree threshold $\tau$ as $acc_\tau = \text{count}(|\theta - \hat{\theta}| < \tau )/n$, following prior works \citep{shen2021gradient, shen2023auxiliary, shen2024task}, where $n$ is the number of test cases; $\theta$ and $\hat{\theta}$ represent the ground truth and the predicted steer angle, respectively, for $\tau\in\mathcal{T}=\{1.5, 3.0, 7.5, 15.0\}$. Then we compute the mean accuracy (mAcc) by averaging $acc_\tau$ across different thresholds. 

\begin{figure}[ht]
    \centering
    \includegraphics[width=0.9\linewidth]{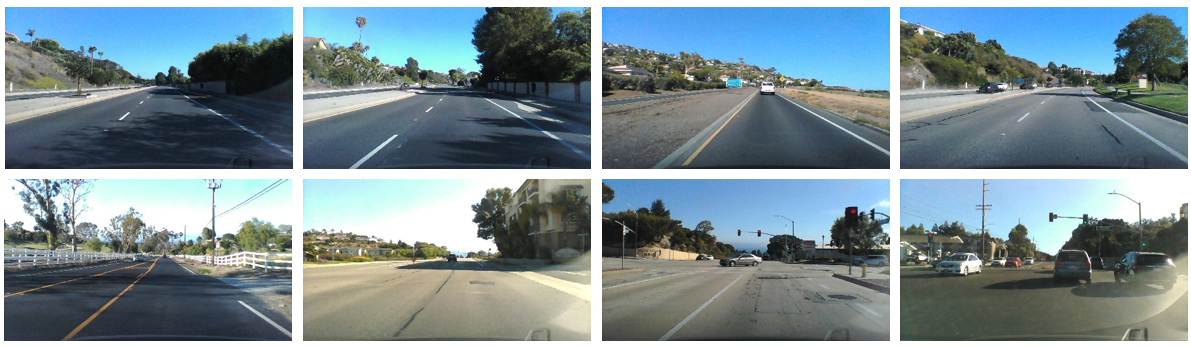}
    % \vspace{-8pt}
    \caption{\textbf{Sample images of the real-world SullyChen dataset.} 
    SullyChen \citep{chen2018collection} is a real-world driving dataset which includes diverse driving scenarios with various road types and conditions.}
    \label{fig:sully}
\end{figure}

The SullyChen \citep{chen2018collection} dataset contains approximately 63,000 images, each with a resolution of 455 $\times$ 256, paired with a corresponding steer angle annotation. We show some sample images in Fig. \ref{fig:sully}. To generate edge maps from RGB images, we employ DexiNed \citep{soria2023dense}. To generate depth maps, we utilize DPT \citep{ranftl2021vision}. Following \citep{shen2023auxiliary}, we use channel-level attention to represent the importance of each modality. For the teacher model $f_p$, we combine the data from different modalities (RGB, depth, and edge) at the channel level and pass them through an Squeeze-and-Excitation (SE) block \citep{hu2018squeeze}, followed by a 1$\times$1 convolution layer to make the channel number to be the same as the main modality RGB. We first train the teacher model offline. Then we use it to guide the target student model $f_t$ training through knowledge distillation, while the RGB images for $f_t$ training has domain shifts compared to the ones used for $f_p$ training. For the details of domain shift, please refer to Appendix \ref{app:ad_data}. 

\begin{figure}[ht]
    \centering
    \includegraphics[width=0.5\linewidth]{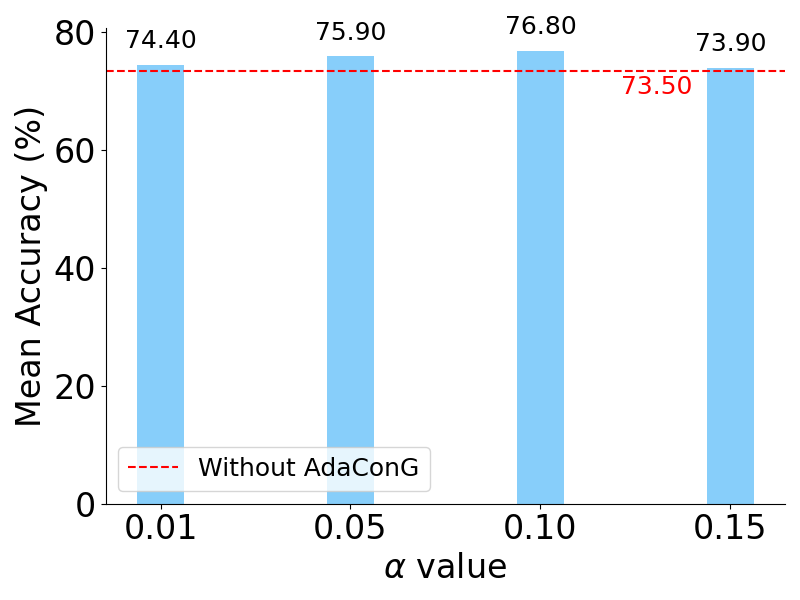}
    \caption{Mean accuracy of KD for steer prediction using \ours~with varying $\alpha$ values.}
    \label{fig:ad_a}
\end{figure}

We compute nonconformity score $s$ as the residuals between the predicted and true steer angles from the calibration set to get the quantile value $q_{1-\alpha}$ with $\alpha=0.1$. A sensitivity analysis for different $\alpha$ values is presented in Fig.~\ref{fig:ad_a}. Then we use it to construct the prediction set $\mathcal{C}(x)$ for a given input RGB image $x$. We define the teacher's uncertainty as the size of the prediction set: $u(x)=|\mathcal{C}(x)|$. The dynamic weight is assigned as $w(x)=1$ if $u(x)<\tau$, otherwise $w(x)=0$, for $\tau \in \mathcal{T}$. We set the coefficients $\lambda_\text{task}$ and $\lambda_\text{guide}$ in Eq. \ref{eq: loss} for all knowledge distillation methods following \citep{shen2023auxiliary}. 

% Here, $\gL_t$ represents the MSE loss, while $\gL_g$ varies based on each knowledge distillation method.

\subsubsection{Data Processing} \label{app:ad_data}
After generating depth and edge maps, we split the dataset into $80\%$ training and $20\%$ testing. Then we train the multi-modal teacher model on the training data offline. After training the teacher model, we introduce domain shift to the training data compared to the teacher's pretraining data. Specifically, we add Gaussian noise with zero mean and a standard deviation of 0.1 to $30\%$ of the RGB images, where the noisy samples are selected uniformly at random across the entire training set to ensure consistent noise distribution. We do not add Gaussian noise to the generated depth and edge maps, as they are not used for the student model. Then we shuffle the training data and randomly split it into a $90\%$ training set $\gD_\text{train}^t$, and a $10\%$ calibration set $\gD_\text{cal}$, ensuring that both sets are drawn from the same underlying distribution. Additionally, the same Gaussian noise is added to $30\%$ of the testing data of RGB images, to form the noisy test set $\gD_\text{test}$, which allows us to evaluate the performance of the models under noise conditions.

\subsubsection{Hyperparameter Sensitivity Analysis}
We conduct a sensitivity analysis for $\alpha$ and present the results in Fig. \ref{fig:ad_a}, illustrating how different values of $\alpha$ affect the student model's mean accuracy (mAcc) on the steer prediction task using KD \cite{hinton2015distilling}. We select $\alpha = 0.10$ that yields slightly better performance.

\subsubsection{Training Details}\label{app:ad_train}
For the experiments, we employ a batch size of 32 and the Adam optimizer \citep{kingma2014adam} with an initial learning rate of $1e{-3}$, and a weight decay of $1e-5$. The model is trained on an Nvidia RTX 3090 GPU with AMD Ryzen 9 5900 CPU and 32 GB RAM for 240 epochs.

\subsection{Disucssions and Future Work} \label{app:limit}
\ours~dynamically modulates the influence of guidance signals, allowing models to reduce reliance on potentially misleading information and thereby enhance learning performance. It also suggests promising directions for future work. Currently, \ours~relies on well-defined ground truth; however, its foundation in conformal prediction allows for a natural extension to settings with ambiguous or imprecise labels \citep{caprio2025conformalized}. By designing nonconformity scores that capture label ambiguity, prediction sets can be constructed to reflect varying degrees of uncertainty. This flexibility enables adaptive weighting even under uncertain supervision. Furthermore, integrating concepts from Imprecise Probabilistic Machine Learning, such as credal sets \citep{caprio2024credal} and imprecise probabilities \citep{dutta2025distribution}, along with strategies from Active Learning and Continual Learning \citep{lu2024ibcl}, could further enhance \ours's capacity to manage complex uncertainty.

% CP relies on the assumption of exchangeability. This generally holds when the calibration data used to conformalize the guidance signal is drawn from the target domain, a condition that is typically easy to satisfy in practice. However, this assumption may be weakened under strong distribution shifts. While we use CP not for strict coverage guarantees but rather to derive a soft uncertainty signal (via prediction set size), extending \ours~to better handle potential violations of exchangeability, such as through covariate-aware CP or adaptive calibration techniques, would be a promising direction for future work.

\end{document}

%% file: math_commands.tex
\usepackage{amsmath,amsfonts,bm}

\def\eqref#1{equation~\ref{#1}}
\def\1{\bm{1}}

\DeclareMathAlphabet{\mathsfit}{\encodingdefault}{\sfdefault}{m}{sl}
\SetMathAlphabet{\mathsfit}{bold}{\encodingdefault}{\sfdefault}{bx}{n}

\def\gA{{\mathcal{A}}}

\def\gC{{\mathcal{C}}}
\def\gD{{\mathcal{D}}}

\def\gL{{\mathcal{L}}}

\def\gP{{\mathcal{P}}}

\def\gR{{\mathcal{R}}}
\def\gS{{\mathcal{S}}}

\newcommand{\E}{\mathbb{E}}

%% file: main.bbl
\begin{thebibliography}{75}
\providecommand{\natexlab}[1]{#1}
\providecommand{\url}[1]{\texttt{#1}}
\expandafter\ifx\csname urlstyle\endcsname\relax
  \providecommand{\doi}[1]{doi: #1}\else
  \providecommand{\doi}{doi: \begingroup \urlstyle{rm}\Url}\fi

\bibitem[Angelopoulos et~al.(2020)Angelopoulos, Bates, Malik, and Jordan]{angelopoulos2020uncertainty}
Anastasios Angelopoulos, Stephen Bates, Jitendra Malik, and Michael~I Jordan.
\newblock Uncertainty sets for image classifiers using conformal prediction.
\newblock \emph{arXiv preprint arXiv:2009.14193}, 2020.

\bibitem[Angelopoulos and Bates(2021)]{angelopoulos2021gentle}
Anastasios~N Angelopoulos and Stephen Bates.
\newblock A gentle introduction to conformal prediction and distribution-free uncertainty quantification.
\newblock \emph{arXiv preprint arXiv:2107.07511}, 2021.

\bibitem[Bhaskar et~al.(2024{\natexlab{a}})Bhaskar, Liu, Sharma, Shi, and Tokekar]{bhaskar2024lava}
Amisha Bhaskar, Rui Liu, Vishnu~D Sharma, Guangyao Shi, and Pratap Tokekar.
\newblock Lava: Long-horizon visual action based food acquisition.
\newblock In \emph{2024 IEEE/RSJ International Conference on Intelligent Robots and Systems (IROS)}, pages 8929--8935. IEEE, 2024{\natexlab{a}}.

\bibitem[Bhaskar et~al.(2024{\natexlab{b}})Bhaskar, Mahammad, Jadhav, and Tokekar]{bhaskar2024planrl}
Amisha Bhaskar, Zahiruddin Mahammad, Sachin~R Jadhav, and Pratap Tokekar.
\newblock Planrl: A motion planning and imitation learning framework to bootstrap reinforcement learning.
\newblock \emph{arXiv preprint arXiv:2408.04054}, 2024{\natexlab{b}}.

\bibitem[Bojarski(2016)]{bojarski2016end}
Mariusz Bojarski.
\newblock End to end learning for self-driving cars.
\newblock \emph{arXiv preprint arXiv:1604.07316}, 2016.

\bibitem[Caprio et~al.(2024)Caprio, Dutta, Jang, Lin, Ivanov, Sokolsky, and Lee]{caprio2024credal}
Michele Caprio, Souradeep Dutta, Kuk~Jin Jang, Vivian Lin, Radoslav Ivanov, Oleg Sokolsky, and Insup Lee.
\newblock Credal bayesian deep learning.
\newblock \emph{Transactions on Machine Learning Research}, 2024.
\newblock ISSN 2835-8856.
\newblock URL \url{https://openreview.net/forum?id=4NHF9AC5ui}.

\bibitem[Caprio et~al.(2025)Caprio, Stutz, Li, and Doucet]{caprio2025conformalized}
Michele Caprio, David Stutz, Shuo Li, and Arnaud Doucet.
\newblock Conformalized credal regions for classification with ambiguous ground truth.
\newblock \emph{Transactions on Machine Learning Research}, 2025.
\newblock ISSN 2835-8856.
\newblock URL \url{https://openreview.net/forum?id=L7sQ8CW2FY}.

\bibitem[Chen(2018)]{chen2018collection}
Sully Chen.
\newblock A collection of labeled car driving datasets.
\newblock \emph{Collection of labeled car driving datasets}, 2018.

\bibitem[Chevalier-Boisvert et~al.(2024)Chevalier-Boisvert, Dai, Towers, Perez-Vicente, Willems, Lahlou, Pal, Castro, and Terry]{chevalier2024minigrid}
Maxime Chevalier-Boisvert, Bolun Dai, Mark Towers, Rodrigo Perez-Vicente, Lucas Willems, Salem Lahlou, Suman Pal, Pablo~Samuel Castro, and Jordan Terry.
\newblock Minigrid \& miniworld: Modular \& customizable reinforcement learning environments for goal-oriented tasks.
\newblock \emph{Advances in Neural Information Processing Systems}, 36, 2024.

\bibitem[Coates et~al.(2011)Coates, Ng, and Lee]{coates2011analysis}
Adam Coates, Andrew Ng, and Honglak Lee.
\newblock An analysis of single-layer networks in unsupervised feature learning.
\newblock In \emph{Proceedings of the fourteenth international conference on artificial intelligence and statistics}, pages 215--223. JMLR Workshop and Conference Proceedings, 2011.

\bibitem[Dutta et~al.(2025)Dutta, Caprio, Lin, Cleaveland, Jang, Ruchkin, Sokolsky, and Lee]{dutta2025distribution}
Souradeep Dutta, Michele Caprio, Vivian Lin, Matthew Cleaveland, Kuk~Jin Jang, Ivan Ruchkin, Oleg Sokolsky, and Insup Lee.
\newblock Distributionally robust statistical verification with imprecise neural networks, 2025.
\newblock URL \url{https://arxiv.org/abs/2308.14815}.

\bibitem[Edupuganti et~al.(2020)Edupuganti, Mardani, Vasanawala, and Pauly]{edupuganti2020uncertainty}
Vineet Edupuganti, Morteza Mardani, Shreyas Vasanawala, and John Pauly.
\newblock Uncertainty quantification in deep mri reconstruction.
\newblock \emph{IEEE Transactions on Medical Imaging}, 40\penalty0 (1):\penalty0 239--250, 2020.

\bibitem[Gal and Ghahramani(2016)]{gal2016dropout}
Yarin Gal and Zoubin Ghahramani.
\newblock Dropout as a bayesian approximation: Representing model uncertainty in deep learning.
\newblock In \emph{international conference on machine learning}, pages 1050--1059. PMLR, 2016.

\bibitem[Gao and Zhang(2021)]{gao2021bayesian}
Peng Gao and Hao Zhang.
\newblock Bayesian deep graph matching for correspondence identification in collaborative perception.
\newblock In \emph{Robotics Science and Systems (RSS)}, 2021.

\bibitem[Gao et~al.(2023)Gao, Zhu, and Zhang]{gao2023uncertainty}
Peng Gao, Qingzhao Zhu, and Hao Zhang.
\newblock Uncertainty-aware correspondence identification for collaborative perception.
\newblock \emph{Autonomous Robots}, 47\penalty0 (5):\penalty0 635--648, 2023.

\bibitem[Gibbs and Candès(2021)]{gibbs2021adaptive}
Isaac Gibbs and Emmanuel Candès.
\newblock Adaptive conformal inference under distribution shift, 2021.
\newblock URL \url{https://arxiv.org/abs/2106.00170}.

\bibitem[Gu et~al.(2023)Gu, Dong, Wei, and Huang]{gu2023knowledge}
Yuxian Gu, Li~Dong, Furu Wei, and Minlie Huang.
\newblock Knowledge distillation of large language models.
\newblock \emph{arXiv preprint arXiv:2306.08543}, 2023.

\bibitem[Haarnoja et~al.(2018)Haarnoja, Zhou, Abbeel, and Levine]{haarnoja2018soft}
Tuomas Haarnoja, Aurick Zhou, Pieter Abbeel, and Sergey Levine.
\newblock Soft actor-critic: Off-policy maximum entropy deep reinforcement learning with a stochastic actor.
\newblock In \emph{International conference on machine learning}, pages 1861--1870. PMLR, 2018.

\bibitem[He et~al.(2016)He, Zhang, Ren, and Sun]{he2016deep}
Kaiming He, Xiangyu Zhang, Shaoqing Ren, and Jian Sun.
\newblock Deep residual learning for image recognition.
\newblock In \emph{Proceedings of the IEEE conference on computer vision and pattern recognition}, pages 770--778, 2016.

\bibitem[Hinton(2015)]{hinton2015distilling}
Geoffrey Hinton.
\newblock Distilling the knowledge in a neural network.
\newblock \emph{arXiv preprint arXiv:1503.02531}, 2015.

\bibitem[Hu et~al.(2023)Hu, Mirchandani, and Sadigh]{hu2023imitation}
Hengyuan Hu, Suvir Mirchandani, and Dorsa Sadigh.
\newblock Imitation bootstrapped reinforcement learning.
\newblock \emph{arXiv preprint arXiv:2311.02198}, 2023.

\bibitem[Hu et~al.(2018)Hu, Shen, and Sun]{hu2018squeeze}
Jie Hu, Li~Shen, and Gang Sun.
\newblock Squeeze-and-excitation networks.
\newblock In \emph{Proceedings of the IEEE conference on computer vision and pattern recognition}, pages 7132--7141, 2018.

\bibitem[Huo et~al.(2024)Huo, Xu, Guo, Wang, and Guo]{huo2024c2kd}
Fushuo Huo, Wenchao Xu, Jingcai Guo, Haozhao Wang, and Song Guo.
\newblock C2kd: Bridging the modality gap for cross-modal knowledge distillation.
\newblock In \emph{Proceedings of the IEEE/CVF Conference on Computer Vision and Pattern Recognition}, pages 16006--16015, 2024.

\bibitem[Jin et~al.(2023)Jin, Wang, and Lin]{jin2023multi}
Ying Jin, Jiaqi Wang, and Dahua Lin.
\newblock Multi-level logit distillation.
\newblock In \emph{Proceedings of the IEEE/CVF Conference on Computer Vision and Pattern Recognition}, pages 24276--24285, 2023.

\bibitem[Kage et~al.(2024)Kage, Rothenberger, Andreadis, and Diochnos]{kage2024review}
Patrick Kage, Jay~C Rothenberger, Pavlos Andreadis, and Dimitrios~I Diochnos.
\newblock A review of pseudo-labeling for computer vision.
\newblock \emph{arXiv preprint arXiv:2408.07221}, 2024.

\bibitem[Karimi and Samavi(2023)]{karimi2023quantifying}
Hamed Karimi and Reza Samavi.
\newblock Quantifying deep learning model uncertainty in conformal prediction.
\newblock In \emph{Proceedings of the AAAI Symposium Series}, volume~1, pages 142--148, 2023.

\bibitem[Kim et~al.(2018)Kim, Park, and Kwak]{kim2018paraphrasing}
Jangho Kim, SeongUk Park, and Nojun Kwak.
\newblock Paraphrasing complex network: Network compression via factor transfer.
\newblock \emph{Advances in neural information processing systems}, 31, 2018.

\bibitem[Kingma(2014)]{kingma2014adam}
Diederik~P Kingma.
\newblock Adam: A method for stochastic optimization.
\newblock \emph{arXiv preprint arXiv:1412.6980}, 2014.

\bibitem[Krizhevsky et~al.(2009)Krizhevsky, Hinton, et~al.]{krizhevsky2009learning}
Alex Krizhevsky, Geoffrey Hinton, et~al.
\newblock Learning multiple layers of features from tiny images.
\newblock 2009.

\bibitem[Kwon et~al.(2020)Kwon, Won, Kim, and Paik]{kwon2020uncertainty}
Yongchan Kwon, Joong-Ho Won, Beom~Joon Kim, and Myunghee~Cho Paik.
\newblock Uncertainty quantification using bayesian neural networks in classification: Application to biomedical image segmentation.
\newblock \emph{Computational Statistics \& Data Analysis}, 142:\penalty0 106816, 2020.

\bibitem[Liu et~al.(2023)Liu, Shi, and Tokekar]{liu2023data}
Rui Liu, Guangyao Shi, and Pratap Tokekar.
\newblock Data-driven distributionally robust optimal control with state-dependent noise.
\newblock In \emph{2023 IEEE/RSJ International Conference on Intelligent Robots and Systems (IROS)}, pages 9986--9991. IEEE, 2023.

\bibitem[Liu et~al.(2024{\natexlab{a}})Liu, Bhaskar, and Tokekar]{liu2024adaptive}
Rui Liu, Amisha Bhaskar, and Pratap Tokekar.
\newblock Adaptive visual imitation learning for robotic assisted feeding across varied bowl configurations and food types.
\newblock \emph{arXiv preprint arXiv:2403.12891}, 2024{\natexlab{a}}.

\bibitem[Liu et~al.(2024{\natexlab{b}})Liu, Gupta, Noorani, and Tokekar]{liu2024towards}
Rui Liu, Anish Gupta, Erfaun Noorani, and Pratap Tokekar.
\newblock Towards efficient risk-sensitive policy gradient: An iteration complexity analysis.
\newblock \emph{arXiv preprint arXiv:2403.08955}, 2024{\natexlab{b}}.

\bibitem[Liu et~al.(2025{\natexlab{a}})Liu, Mahammad, Bhaskar, and Tokekar]{liu2025imrl}
Rui Liu, Zahiruddin Mahammad, Amisha Bhaskar, and Pratap Tokekar.
\newblock Imrl: Integrating visual, physical, temporal, and geometric representations for enhanced food acquisition.
\newblock In \emph{2025 IEEE International Conference on Robotics and Automation (ICRA)}, pages 741--747. IEEE, 2025{\natexlab{a}}.

\bibitem[Liu et~al.(2025{\natexlab{b}})Liu, Shen, Gao, Tokekar, and Lin]{liu2025caml}
Rui Liu, Yu~Shen, Peng Gao, Pratap Tokekar, and Ming Lin.
\newblock Caml: Collaborative auxiliary modality learning for multi-agent systems.
\newblock \emph{arXiv preprint arXiv:2502.17821}, 2025{\natexlab{b}}.

\bibitem[Liu et~al.(2025{\natexlab{c}})Liu, Wang, Gao, Shen, Tokekar, and Lin]{liu2025mmcd}
Rui Liu, Zikang Wang, Peng Gao, Yu~Shen, Pratap Tokekar, and Ming Lin.
\newblock Mmcd: Multi-modal collaborative decision-making for connected autonomy with knowledge distillation.
\newblock \emph{arXiv preprint arXiv:2509.18198}, 2025{\natexlab{c}}.

\bibitem[Lu et~al.(2022)Lu, Angelopoulos, and Pomerantz]{lu2022improving}
Charles Lu, Anastasios~N Angelopoulos, and Stuart Pomerantz.
\newblock Improving trustworthiness of ai disease severity rating in medical imaging with ordinal conformal prediction sets.
\newblock In \emph{International Conference on Medical Image Computing and Computer-Assisted Intervention}, pages 545--554. Springer, 2022.

\bibitem[Lu et~al.(2024)Lu, Caprio, Eaton, and Lee]{lu2024ibcl}
Pengyuan Lu, Michele Caprio, Eric Eaton, and Insup Lee.
\newblock Ibcl: Zero-shot model generation under stability-plasticity trade-offs, 2024.
\newblock URL \url{https://arxiv.org/abs/2305.14782}.

\bibitem[Ma et~al.(2018)Ma, Zhang, Zheng, and Sun]{ma2018shufflenet}
Ningning Ma, Xiangyu Zhang, Hai-Tao Zheng, and Jian Sun.
\newblock Shufflenet v2: Practical guidelines for efficient cnn architecture design.
\newblock In \emph{Proceedings of the European conference on computer vision (ECCV)}, pages 116--131, 2018.

\bibitem[Mossina et~al.(2024)Mossina, Dalmau, and And{\'e}ol]{mossina2024conformal}
Luca Mossina, Joseba Dalmau, and L{\'e}o And{\'e}ol.
\newblock Conformal semantic image segmentation: Post-hoc quantification of predictive uncertainty.
\newblock In \emph{Proceedings of the IEEE/CVF Conference on Computer Vision and Pattern Recognition}, pages 3574--3584, 2024.

\bibitem[Namdari and Li(2019)]{namdari2019review}
Alireza Namdari and Zhaojun Li.
\newblock A review of entropy measures for uncertainty quantification of stochastic processes.
\newblock \emph{Advances in Mechanical Engineering}, 11\penalty0 (6):\penalty0 1687814019857350, 2019.

\bibitem[Passalis and Tefas(2018)]{passalis2018learning}
Nikolaos Passalis and Anastasios Tefas.
\newblock Learning deep representations with probabilistic knowledge transfer.
\newblock In \emph{Proceedings of the European Conference on Computer Vision (ECCV)}, pages 268--284, 2018.

\bibitem[Pearce et~al.(2021)Pearce, Brintrup, and Zhu]{pearce2021understanding}
Tim Pearce, Alexandra Brintrup, and Jun Zhu.
\newblock Understanding softmax confidence and uncertainty.
\newblock \emph{arXiv preprint arXiv:2106.04972}, 2021.

\bibitem[Ranftl et~al.(2021)Ranftl, Bochkovskiy, and Koltun]{ranftl2021vision}
Ren{\'e} Ranftl, Alexey Bochkovskiy, and Vladlen Koltun.
\newblock Vision transformers for dense prediction.
\newblock In \emph{Proceedings of the IEEE/CVF international conference on computer vision}, pages 12179--12188, 2021.

\bibitem[Romero et~al.(2014)Romero, Ballas, Kahou, Chassang, Gatta, and Bengio]{romero2014fitnets}
Adriana Romero, Nicolas Ballas, Samira~Ebrahimi Kahou, Antoine Chassang, Carlo Gatta, and Yoshua Bengio.
\newblock Fitnets: Hints for thin deep nets.
\newblock \emph{arXiv preprint arXiv:1412.6550}, 2014.

\bibitem[Sandler et~al.(2018)Sandler, Howard, Zhu, Zhmoginov, and Chen]{sandler2018mobilenetv2}
Mark Sandler, Andrew Howard, Menglong Zhu, Andrey Zhmoginov, and Liang-Chieh Chen.
\newblock Mobilenetv2: Inverted residuals and linear bottlenecks.
\newblock In \emph{Proceedings of the IEEE conference on computer vision and pattern recognition}, pages 4510--4520, 2018.

\bibitem[Scherer et~al.(2022)Scherer, Sch{\"o}n, and Lienhart]{scherer2022pseudo}
Sebastian Scherer, Robin Sch{\"o}n, and Rainer Lienhart.
\newblock Pseudo-label noise suppression techniques for semi-supervised semantic segmentation.
\newblock \emph{arXiv preprint arXiv:2210.10426}, 2022.

\bibitem[Shafer and Vovk(2008)]{shafer2008tutorial}
Glenn Shafer and Vladimir Vovk.
\newblock A tutorial on conformal prediction.
\newblock \emph{Journal of Machine Learning Research}, 9\penalty0 (3), 2008.

\bibitem[Shen et~al.(2021)Shen, Zheng, Shu, Li, Goldstein, and Lin]{shen2021gradient}
Yu~Shen, Laura Zheng, Manli Shu, Weizi Li, Tom Goldstein, and Ming Lin.
\newblock Gradient-free adversarial training against image corruption for learning-based steering.
\newblock \emph{Advances in Neural Information Processing Systems}, 34:\penalty0 26250--26263, 2021.

\bibitem[Shen et~al.(2023)Shen, Wang, Gao, and Lin]{shen2023auxiliary}
Yu~Shen, Xijun Wang, Peng Gao, and Ming Lin.
\newblock Auxiliary modality learning with generalized curriculum distillation.
\newblock In \emph{International Conference on Machine Learning}, pages 31057--31076. PMLR, 2023.

\bibitem[Shen et~al.(2024)Shen, Zheng, Zhou, and Lin]{shen2024task}
Yu~Shen, Laura Zheng, Tianyi Zhou, and C~Lin.
\newblock Task-driven domain-agnostic learning with information bottleneck for autonomous steering.
\newblock In \emph{2024 IEEE International Conference on Robotics and Automation (ICRA)}, pages 6858--6865. IEEE, 2024.

\bibitem[Simonyan(2014)]{simonyan2014very}
Karen Simonyan.
\newblock Very deep convolutional networks for large-scale image recognition.
\newblock \emph{arXiv preprint arXiv:1409.1556}, 2014.

\bibitem[Sohn et~al.(2020)Sohn, Berthelot, Carlini, Zhang, Zhang, Raffel, Cubuk, Kurakin, and Li]{sohn2020fixmatch}
Kihyuk Sohn, David Berthelot, Nicholas Carlini, Zizhao Zhang, Han Zhang, Colin~A Raffel, Ekin~Dogus Cubuk, Alexey Kurakin, and Chun-Liang Li.
\newblock Fixmatch: Simplifying semi-supervised learning with consistency and confidence.
\newblock \emph{Advances in neural information processing systems}, 33:\penalty0 596--608, 2020.

\bibitem[Soria et~al.(2023)Soria, Sappa, Humanante, and Akbarinia]{soria2023dense}
Xavier Soria, Angel Sappa, Patricio Humanante, and Arash Akbarinia.
\newblock Dense extreme inception network for edge detection.
\newblock \emph{Pattern Recognition}, 139:\penalty0 109461, 2023.

\bibitem[Su et~al.(2025)Su, Tseng, Pu, Zhao, Chen, and Lee]{su2025eakdentropybasedadaptiveknowledge}
Chi-Ping Su, Ching-Hsun Tseng, Bin Pu, Lei Zhao, Zhuangzhuang Chen, and Shin-Jye Lee.
\newblock Ea-kd: Entropy-based adaptive knowledge distillation, 2025.
\newblock URL \url{https://arxiv.org/abs/2311.13621}.

\bibitem[Sun et~al.(2024)Sun, Ren, Li, Wang, and Cao]{sun2024logit}
Shangquan Sun, Wenqi Ren, Jingzhi Li, Rui Wang, and Xiaochun Cao.
\newblock Logit standardization in knowledge distillation.
\newblock In \emph{Proceedings of the IEEE/CVF Conference on Computer Vision and Pattern Recognition}, pages 15731--15740, 2024.

\bibitem[Sutskever et~al.(2013)Sutskever, Martens, Dahl, and Hinton]{sutskever2013importance}
Ilya Sutskever, James Martens, George Dahl, and Geoffrey Hinton.
\newblock On the importance of initialization and momentum in deep learning.
\newblock In \emph{International conference on machine learning}, pages 1139--1147. PMLR, 2013.

\bibitem[Tian et~al.(2019)Tian, Krishnan, and Isola]{tian2019contrastive}
Yonglong Tian, Dilip Krishnan, and Phillip Isola.
\newblock Contrastive representation distillation.
\newblock \emph{arXiv preprint arXiv:1910.10699}, 2019.

\bibitem[Tibshirani et~al.(2019)Tibshirani, Foygel~Barber, Candes, and Ramdas]{tibshirani2019conformal}
Ryan~J Tibshirani, Rina Foygel~Barber, Emmanuel Candes, and Aaditya Ramdas.
\newblock Conformal prediction under covariate shift.
\newblock \emph{Advances in neural information processing systems}, 32, 2019.

\bibitem[Torabi et~al.(2018)Torabi, Warnell, and Stone]{torabi2018behavioral}
Faraz Torabi, Garrett Warnell, and Peter Stone.
\newblock Behavioral cloning from observation.
\newblock \emph{arXiv preprint arXiv:1805.01954}, 2018.

\bibitem[Vovk et~al.(2016)Vovk, Fedorova, Nouretdinov, and Gammerman]{vovk2016criteria}
Vladimir Vovk, Valentina Fedorova, Ilia Nouretdinov, and Alexander Gammerman.
\newblock Criteria of efficiency for conformal prediction.
\newblock In \emph{Conformal and Probabilistic Prediction with Applications: 5th International Symposium, COPA 2016, Madrid, Spain, April 20-22, 2016, Proceedings 5}, pages 23--39. Springer, 2016.

\bibitem[Vovk et~al.(2020)Vovk, Petej, Toccaceli, Gammerman, Ahlberg, and Carlsson]{vovk2020conformal}
Vladimir Vovk, Ivan Petej, Paolo Toccaceli, Alexander Gammerman, Ernst Ahlberg, and Lars Carlsson.
\newblock Conformal calibrators.
\newblock In \emph{conformal and probabilistic prediction and applications}, pages 84--99. PMLR, 2020.

\bibitem[Wang et~al.(2023)Wang, Yan, Zhang, Wei, Li, and Li]{wang2023prototype}
Shuai Wang, Zipei Yan, Daoan Zhang, Haining Wei, Zhongsen Li, and Rui Li.
\newblock Prototype knowledge distillation for medical segmentation with missing modality.
\newblock In \emph{ICASSP 2023-2023 IEEE International Conference on Acoustics, Speech and Signal Processing (ICASSP)}, pages 1--5. IEEE, 2023.

\bibitem[Wang et~al.(2020)Wang, Zhang, Tian, Zhong, Shi, Zhang, and He]{wang2020double}
Yixin Wang, Yao Zhang, Jiang Tian, Cheng Zhong, Zhongchao Shi, Yang Zhang, and Zhiqiang He.
\newblock Double-uncertainty weighted method for semi-supervised learning.
\newblock In \emph{Medical Image Computing and Computer Assisted Intervention--MICCAI 2020: 23rd International Conference, Lima, Peru, October 4--8, 2020, Proceedings, Part I 23}, pages 542--551. Springer, 2020.

\bibitem[Xia et~al.(2023)Xia, Wang, Zhou, Hua, and Tang]{xia2023learning}
Kun Xia, Le~Wang, Sanping Zhou, Gang Hua, and Wei Tang.
\newblock Learning from noisy pseudo labels for semi-supervised temporal action localization.
\newblock In \emph{Proceedings of the IEEE/CVF International Conference on Computer Vision}, pages 10160--10169, 2023.

\bibitem[Xie et~al.(2020)Xie, Dai, Hovy, Luong, and Le]{xie2020unsupervised}
Qizhe Xie, Zihang Dai, Eduard Hovy, Thang Luong, and Quoc Le.
\newblock Unsupervised data augmentation for consistency training.
\newblock \emph{Advances in neural information processing systems}, 33:\penalty0 6256--6268, 2020.

\bibitem[Xue et~al.(2022)Xue, Gao, Ren, and Zhao]{xue2022modality}
Zihui Xue, Zhengqi Gao, Sucheng Ren, and Hang Zhao.
\newblock The modality focusing hypothesis: Towards understanding crossmodal knowledge distillation.
\newblock \emph{arXiv preprint arXiv:2206.06487}, 2022.

\bibitem[Yu et~al.(2024)Yu, Mishra, Koppel, Busart, Narayan, Manocha, Bedi, and Tokekar]{yu2024beyond}
Peihong Yu, Manav Mishra, Alec Koppel, Carl Busart, Priya Narayan, Dinesh Manocha, Amrit Bedi, and Pratap Tokekar.
\newblock Beyond joint demonstrations: Personalized expert guidance for efficient multi-agent reinforcement learning.
\newblock \emph{arXiv preprint arXiv:2403.08936}, 2024.

\bibitem[Zagoruyko(2016)]{zagoruyko2016wide}
Sergey Zagoruyko.
\newblock Wide residual networks.
\newblock \emph{arXiv preprint arXiv:1605.07146}, 2016.

\bibitem[Zagoruyko and Komodakis(2016)]{zagoruyko2016paying}
Sergey Zagoruyko and Nikos Komodakis.
\newblock Paying more attention to attention: Improving the performance of convolutional neural networks via attention transfer.
\newblock \emph{arXiv preprint arXiv:1612.03928}, 2016.

\bibitem[Zhang et~al.(2021)Zhang, Wang, Hou, Wu, Wang, Okumura, and Shinozaki]{zhang2021flexmatch}
Bowen Zhang, Yidong Wang, Wenxin Hou, Hao Wu, Jindong Wang, Manabu Okumura, and Takahiro Shinozaki.
\newblock Flexmatch: Boosting semi-supervised learning with curriculum pseudo labeling.
\newblock \emph{Advances in neural information processing systems}, 34:\penalty0 18408--18419, 2021.

\bibitem[Zhang et~al.(2024)Zhang, Shen, Liu, Liu, Bendersky, Najork, and Zhang]{zhang2024knowledge}
Rongzhi Zhang, Jiaming Shen, Tianqi Liu, Jialu Liu, Michael Bendersky, Marc Najork, and Chao Zhang.
\newblock Knowledge distillation with perturbed loss: From a vanilla teacher to a proxy teacher.
\newblock In \emph{Proceedings of the 30th ACM SIGKDD Conference on Knowledge Discovery and Data Mining}, pages 4278--4289, 2024.

\bibitem[Zhang et~al.(2018)Zhang, Zhou, Lin, and Sun]{zhang2018shufflenet}
Xiangyu Zhang, Xinyu Zhou, Mengxiao Lin, and Jian Sun.
\newblock Shufflenet: An extremely efficient convolutional neural network for mobile devices.
\newblock In \emph{Proceedings of the IEEE conference on computer vision and pattern recognition}, pages 6848--6856, 2018.

\bibitem[Zhao et~al.(2024)Zhao, Wang, Xiao, Gao, and Gao]{zhao2024leveraging}
Rui Zhao, Kui Wang, Yang Xiao, Fei Gao, and Zhenhai Gao.
\newblock Leveraging monte carlo dropout for uncertainty quantification in real-time object detection of autonomous vehicles.
\newblock \emph{IEEE Access}, 2024.

\bibitem[Zhou et~al.(2025)Zhou, Zhang, and Luo]{zhou2025computation}
Hao Zhou, Yanze Zhang, and Wenhao Luo.
\newblock Computationally and sample efficient safe reinforcement learning using adaptive conformal prediction, 2025.
\newblock URL \url{https://arxiv.org/abs/2503.17678}.

\end{thebibliography}
